\pdfoutput=1
\RequirePackage{fix-cm}
\documentclass[twocolumn]{svjour3}          
\smartqed  
\usepackage{booktabs}
\usepackage{amsmath}
\usepackage{amssymb}
\usepackage{mathtools}
\usepackage{xspace}
\usepackage{hyperref}
\usepackage{graphicx}
\usepackage{caption}
\usepackage{subcaption}
\usepackage{multirow}
\usepackage{color, colortbl}
\usepackage{fancyvrb}
\usepackage{appendix}
\usepackage{natbib}

\newcommand{\eg}{\emph{e.g.,}\xspace}

\newcommand{\ie}{\emph{i.e.,}\xspace}

\newcommand{\overbar}[1]{\mkern 1.5mu\overline{\mkern-1.5mu#1\mkern-1.5mu}\mkern 1.5mu}

\DeclareMathOperator*{\argmax}{arg\,max}

\definecolor{Gray}{gray}{0.9}
\definecolor{pink}{rgb}{1.0, 0.13, 0.32}

\journalname{International Journal of Computer Vision}

\hypersetup{pdfstartview={FitH}}
\hypersetup{pdfborder={0 0 0}}
\hypersetup{pdfpagemode={UseNone}}
\hypersetup{colorlinks=true}
\hypersetup{citecolor=blue}
\hypersetup{linkcolor=blue}

\begin{document}

\title{Pointly-Supervised Action Localization
}


\author{Pascal~Mettes         \and
        Cees~G.~M.~Snoek 
}


\institute{Pascal Mettes \at
              Universiteit van Amsterdam, Amsterdam, the Netherlands\\
              \email{P.S.M.Mettes@uva.nl}           
           \and
           Cees G. M. Snoek \at
              Universiteit van Amsterdam, Amsterdam, the Netherlands\\
              \email{cgmsnoek@uva.nl}
}

\date{Received: date / Accepted: date}

\maketitle

\begin{abstract}
This paper strives for spatio-temporal localization of human actions in videos. In the literature, the consensus is to achieve localization by training on bounding box annotations provided for each frame of each training video. As annotating boxes in video is expensive, cumbersome and error-prone, we propose to bypass box-supervision. Instead, we introduce action localization based on point-supervision.
We start from unsupervised spatio-temporal proposals, which provide a set of candidate regions in videos. While normally used exclusively for inference, we show spatio-temporal proposals can also be leveraged during training when guided by a sparse set of point annotations. We introduce an overlap measure between points and spatio-temporal proposals and incorporate them all into a new objective of a Multiple Instance Learning optimization. During inference, we introduce pseudo-points, visual cues from videos, that automatically guide the selection of spatio-temporal proposals. We outline five spatial and one temporal pseudo-point, as well as a measure to best leverage pseudo-points at test time. Experimental evaluation on three action localization datasets shows our pointly-supervised approach (\textit{i}) is as effective as traditional box-supervision at a fraction of the annotation cost, (\textit{ii}) is robust to sparse and noisy point annotations, (\textit{iii}) benefits from pseudo-points during inference, and (\textit{iv}) outperforms recent weakly-supervised alternatives.  This leads us to conclude that points provide a viable alternative to boxes for action localization.
\end{abstract}

\section{Introduction}

\begin{figure*}[t]
\includegraphics[width=\textwidth]{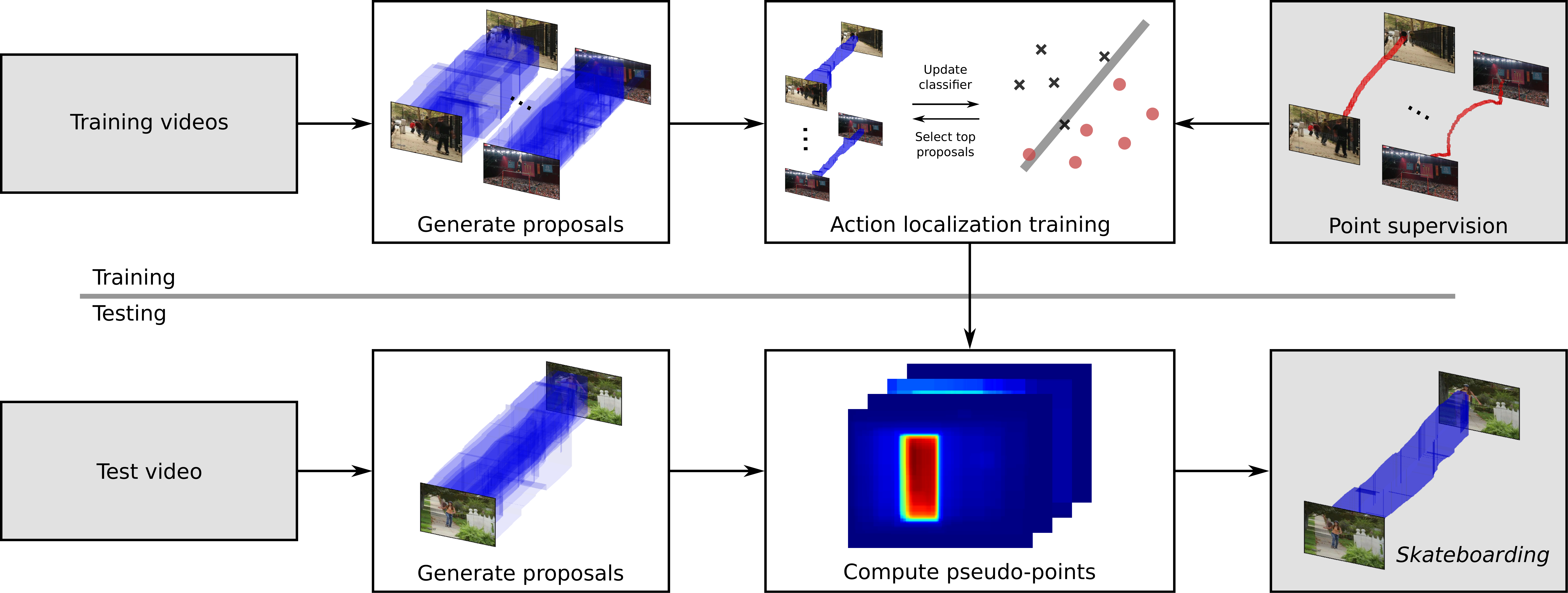}
\caption{Pointly-supervised action localization using spatio-temporal proposals and pseudo-points. During training, we start from point-supervision for each video. Our overlap measure computes the match between each proposal and the point annotations. We iteratively refine the proposal selection by extending the max-margin Multiple Instance Learning formulation. During inference, we compute pseudo-points for all video frames and use them in conjunction with the learned action model to determine the top proposals per action over all test videos.}
\label{fig:fig1}
\end{figure*}

This paper aims to recognize and localize actions such as \emph{skiing}, \emph{running}, and \emph{getting out of a vehicle} in videos. Action recognition has been a vibrant topic in vision for several decades, resulting in approaches based on local spatio-temporal features~\citep{dollar2005behavior,laptev2005space,wang2009evaluation},  dense trajectories~\citep{jain2013better,wang2013dense} two-stream neural networks~\citep{simonyan2014two,feichtenhofer2016convolutional}, 3D convolutions~\citep{ji20133d,tran2015learning}, and recurrent networks~\citep{donahue2015long,li2018videolstm,srivastava2015unsupervised}. We aim to not only recognize which actions occur in videos, but also discover when and where the actions are present.

Action localization in videos corresponds to finding tubes of consecutive bounding boxes in frames for each action.
Initial work aimed at localizing actions by finding local discriminative parts and generating tubes through linking or sliding windows~\citep{lan2011discriminative,tian2013spatiotemporal,wang2014video}. State-of-the-art localizers classify boxes per frame (or few frames) before linking them into tubes~\citep{gkioxari2015finding,weinzaepfelICCV2015learningToTrack,hou2017tube,kalogeiton2017action}. Regardless the approach, a requirement for all these works is the need for box-supervision per frame of each training video. As annotating boxes in videos is an expensive, cumbersome and error-prone endeavor, we prefer to perform action localization without the need for box supervision.

The first contribution of this paper is to localize actions in videos with the aid of point-supervision. For pointly-supervised action localization, we start from (unsupervised) spatio-temporal proposals. Spatio-temporal proposals reduce the search space of actions in videos to a few hundred to thousand tubes, where at least one tube matches well with the ground truth action location~\citep{jain2014action,gemert2015apt,jain2017tubelets,oneata2014spatio}. This is typically achieved by clustering local representations such as supervoxels or dense trajectories. In the literature, the use of spatio-temporal proposals is restricted to the inference stage; training of the action localizer that select the best proposal still depends on box-supervision. While the spatio-temporal proposals may be unsupervised, they do not relax the need for box-supervision during the training stage of action localizers. We propose to bypass bounding box annotations by training action localizers on spatio-temporal proposals from \emph{training} videos. We show that training on spatio-temporal proposals guided by point annotations, yields similar action localization performance to their box-supervised alternative at a fraction of the annotation time.

As our second contribution, we propose an overlap measure that matches the centers of spatio-temporal proposals with point annotations. To identify the best proposal to train on, we adopt a Multiple Instance Learning perspective~\citep{andrews2002support}, with the spatio-temporal proposals defining the instances and videos the bags. We employ the max-margin Multiple Instance Learning formulation and extend it to incorporate information from the proposed overlap measure. This results in action localization using video labels and point annotations as the sole action supervision. Our first two contributions were previously presented in the conference version of this paper~\citep{mettes2016spot}. 

For our third contribution we are inspired by~\cite{mettes2017localizing}, who propose to train action localizers with spatio-temporal proposals selected by automatic visual cues. Rather than employing the cues at training time, we prefer to exploit the cues during \emph{inference} and call them pseudo-points. The pseudo-points are used as an unsupervised generalization of point-supervision during the testing stage. The pseudo-points cover cues from training point statistics, person detection \citep{yuCVPR2015fap}, independent motion \citep{jain2014action}, spatio-temporal proposals \citep{gemert2015apt}, center bias \citep{tseng2009quantifying}, and temporal information. To link the point-supervision in training videos to pseudo-points in test videos, we propose a function that both weights and selects pseudo-points based on how well they match with points annotated during training. We use the weighting function to determine which pseudo-points are most effective and how much they should contribute to the selection of spatio-temporal proposals in test videos. A complete overview of our proposed approach is shown in Figure~\ref{fig:fig1}.

The rest of the paper is organized as follows. In Section~\ref{sec:relwork}, we describe related work. Section~\ref{sec:method1} details our algorithm for point-supervision during training. Section~\ref{sec:method2} presents pseudo-points and explains how to leverage them during inference. We detail our experimental setup on UCF Sports \citep{RodriguezCVPR2008}, UCF-101 \citep{soomro2012ucf101} and Hollywood2Tubes \citep{mettes2016spot} in Section~\ref{sec:exp1}. Ablation studies, error diagnosis and comparisons are discussed in Section~\ref{sec:exp2}. We conclude our work in Section~\ref{sec:conclusions}.

\section{Related work}
\label{sec:relwork}

\subsection{Action localization with box-supervision}
The problem of action localization is commonly tackled by supervision with video-level action class labels and frame-level box annotations during training.
Initial approaches do so through figure-centric structures~\citep{lan2011discriminative} and part-based models~\citep{tian2013spatiotemporal,wang2014video}. 
Inspired by the success of object proposals in images~\citep{uijlings2013selective}, several works have investigated spatio-temporal proposals for action localization in videos. Such spatio-temporal proposals are typically generated by grouping supervoxels~\citep{jain2017tubelets,oneata2014spatio,soomroICCV2015actionLocContextWalk} or dense trajectories~\citep{gemert2015apt,marianICCV2015unsupervisedTube}. Spatio-temporal proposals reduce the search space to a few hundred or thousand locations per video. In the literature, the use of spatio-temporal proposals is limited to the testing stage. Training is still performed on features derived from bounding box annotations per frame. In this paper, we extend the use of action proposals to the training stage. We show that proposals provide high quality training examples when leveraging our Multiple Instance Learning variant, guided by point annotations, completely alleviating the need for box annotations.

Recently, a number of works have achieved success in action localization by separating spatial detection from temporal linking~\citep{gkioxari2015finding,weinzaepfelICCV2015learningToTrack}. Such approaches have been further improved with better representations~\citep{peng2016multi,saha2016deep,yang2017spatio}, joint linking algorithms~\citep{singh2017online}, and by classifying a few consecutive frames before linking~\citep{hou2017tube,kalogeiton2017action,saha2017amtnet}. While effective, these approaches have an inherent requirement for box annotations to detect and regress the boxes in video frames. 
We focus on the use of unsupervised spatio-temporal proposals~\citep{jain2014action,gemert2015apt,jain2017tubelets,oneata2014spatio}, and we show how to utilize them during training to bypass the need for box-supervision.
%
%

\subsection{Action localization without box-supervision}
Given the annotation burden for box-supervision in action localization, several works have investigated action localization from weaker supervision signals. Most works focus on localization from video labels only. \cite{siva2011weakly} employ spatio-temporal proposals and optimize for an action localization model through Multiple Instance Learning~\citep{andrews2002support}, where the videos are the bags and the proposals are the instances. We show that Multiple Instance Learning yields suboptimal results for action localization; extending Multiple Instance Learning with point-supervision alleviates this problem.

\cite{chen2015action} also employ spatio-temporal proposals and video labels, but skip the Multiple Instance Learning step. Instead, they train on the most dominant proposal per training video, without knowing whether the proposal fits the action location well. Recent work by \cite{li2018videolstm} achieves action localization from video labels through attention. The action location is determined by a box around the center of attention in each frame, followed by a linking procedure. These approaches provide action localization without box annotations. However, using only the video label restricts the localization performance. We show that point annotations have a direct impact on the performance at the expense of a small additional annotation cost, outperforming approaches using video labels only.

Several recent works have investigated action localization in a zero-shot setting, where no video training examples are provided for test actions. This is typically achieved through semantic word embeddings~\citep{mikolov2013distributed} between actions and objects as found in text corpora. Initial work by \cite{jain2015objects2action} employed spatio-temporal proposals and assigned object classifier scores to each proposal. The object scores are combined with the word embedding scores given an action and the highest scoring proposal is selected for each test video. \cite{mettes2017spatial} perform zero-shot action localization by linking boxes that are scored based on a spatial-aware embedding between actors and objects. \cite{kalogeiton2017joint} perform zero-shot localization through joint localization of actions and objects. \cite{soomro2017unsupervised} aim for unsupervised action localization through discriminative clustering on videos and spatio-temporal action proposal generation with 0-1 Knapsack. Such works are promising but do not perform on the level of (weakly) supervised alternatives, as detailed in our final experiment.

\subsection{Speeding-up box annotations}
Easing the annotation burden of bounding box annotations in videos has been investigated by~\cite{vondrickIJCV2013crowdsourced}. 
They investigate different strategies to annotate boxes in videos, \eg with expert annotators and tracking.
Furthermore, several works have attempted faster ways to annotate boxes, \eg through human verification~\citep{russakovsky2015best,papadopoulos2016we} or by clicking the extremes of objects~\citep{papadopoulos2017extreme}. While such investigations and approaches provide faster alternatives to the costly ImageNet standard for box annotation~\citep{su2012crowdsourcing}, annotating boxes remains a slow and manually expensive endeavor In this work, we avoid box annotations and show that action localization can be done efficiently through simple point annotations.

Several recent works have investigated the merit of point annotations in other visual recognition challenges. \cite{bearman2016s} investigate point-supervision for semantic segmentation in images, which constitutes a fraction of the annotation cost compared to pixel-wise segmentation. In the video domain, \cite{jain2016click} investigate object segmentation based on point clicks. Similar in spirit to our work, \cite{manen2017pathtrack} show the spatio-temporal tracks from consecutive point annotations provide a rich supervision for multiple object tracking in videos. In this work, we investigate the potential of point-supervision for action localization in videos, showing we can reach comparable performance to full box-supervision approaches based on action proposals.

%
%
\section{Point-supervision for training} \label{sec:method1}

For pointly-supervised action localization with spatio-temporal proposals, we start from the hypothesis that the proposals themselves, normally used for testing, can substitute the ground truth box annotations during training without a significant loss in performance.
%
%
%
Our main goal is to mine out of a set of action proposals the best one during training while minimizing the annotation effort.
The first level of supervision constitutes the action class label for the whole video. Given such global labels, we follow the traditional approach of mining the best proposals through Multiple Instance Learning, as introduced for object detection by~\cite{cinbis2017weakly}. In the context of action localization, each video is interpreted as a bag and the proposals in each video are interpreted as its instances. The goal of Multiple Instance Learning is to train a classifier that selects the top proposals and separates proposals from different actions.

Next to the global action class label we leverage easy to obtain annotations within each video: we simply point at the action. Point-supervision allows us to easily exclude those proposals that have no overlap with any annotated point. Nevertheless, there are still many proposals that intersect with at least one point, as points do not uniquely identify each proposal. Therefore, we introduce an overlap measure to associate proposals with points. We also extend the objective of Multiple Instance Learning to include the proposed overlap measure for proposal mining.
%

\subsection{Overlap between proposals and points}
Let us first introduce the following notation.
For a video $V$ of $F_V$ frames, an action tube proposal  $A=\{\text{BB}_i \}_{\text{i}=f}^m$ consists of connected bounding boxes through video frames $(f,...,m)$ where $1 \le f < m \le F_V$.
Let $\overbar{BB_{i}}$ denote the center of a bounding box $i$.
The point supervision $C=\{(c_{i}^{(x)}, c_{i}^{(y)})\}^K$ is a set of $K \le F_V$ sub-sampled video frames where each frame $i$ has a single annotated point $(c_{i}^{(x)}, c_{i}^{(y)})$.
We propose an overlap measure that provides a continuous bounded score based on the match between a proposal and the point annotations.

Our overlap measure, inspired by a mild center-bias in annotators~\citep{tseng2009quantifying}, consists of two terms.
The first term $M(\cdot)$ states how close the center of a bounding box from a proposal is to an annotated point, relative to the bounding box size, within the same frame. This center-bias term normalizes the distance of a point annotation to the center of a bounding box by the distance between the center and closest edge of the bounding box.
For point annotation $(c_{i}^{(x)}, c_{i}^{(y)})$ and for bounding box $BB_{K_i}$ in the same frame, the score is 1 if the box center $\overbar{BB_{K_i}}$ is the same as the point annotation. The score decreases linearly in value as the distance between the point annotation and the box center grows and the score becomes 0 if the point annotation is not contained in $BB_{K_i}$:
\begin{equation}
\resizebox{\hsize}{!}{
$M(A, C)  =   \frac{1}{K} \sum_{i=1}^{K}  \text{max}(0, 1 - \frac{||(c_{i}^{(x)}, c_{i}^{(y)}) - \overbar{BB_{K_i}} ||_2}{ \max\limits_{(u,v) \in e(BB_{K_i})} ||( (u,v) - \overbar{BB_{K_i}}) ||_2}).$
}
\label{eq:overlap1}
\end{equation}
In Equation 1, $(u,v)$ denotes the center point of each of the four edges of box $BB_{K_i}$, given by the function $e(BB_{K_i})$.

\begin{figure}[t]
	\centering
	\begin{subfigure}{0.32\linewidth}
		\centering
		\includegraphics[width=\textwidth]{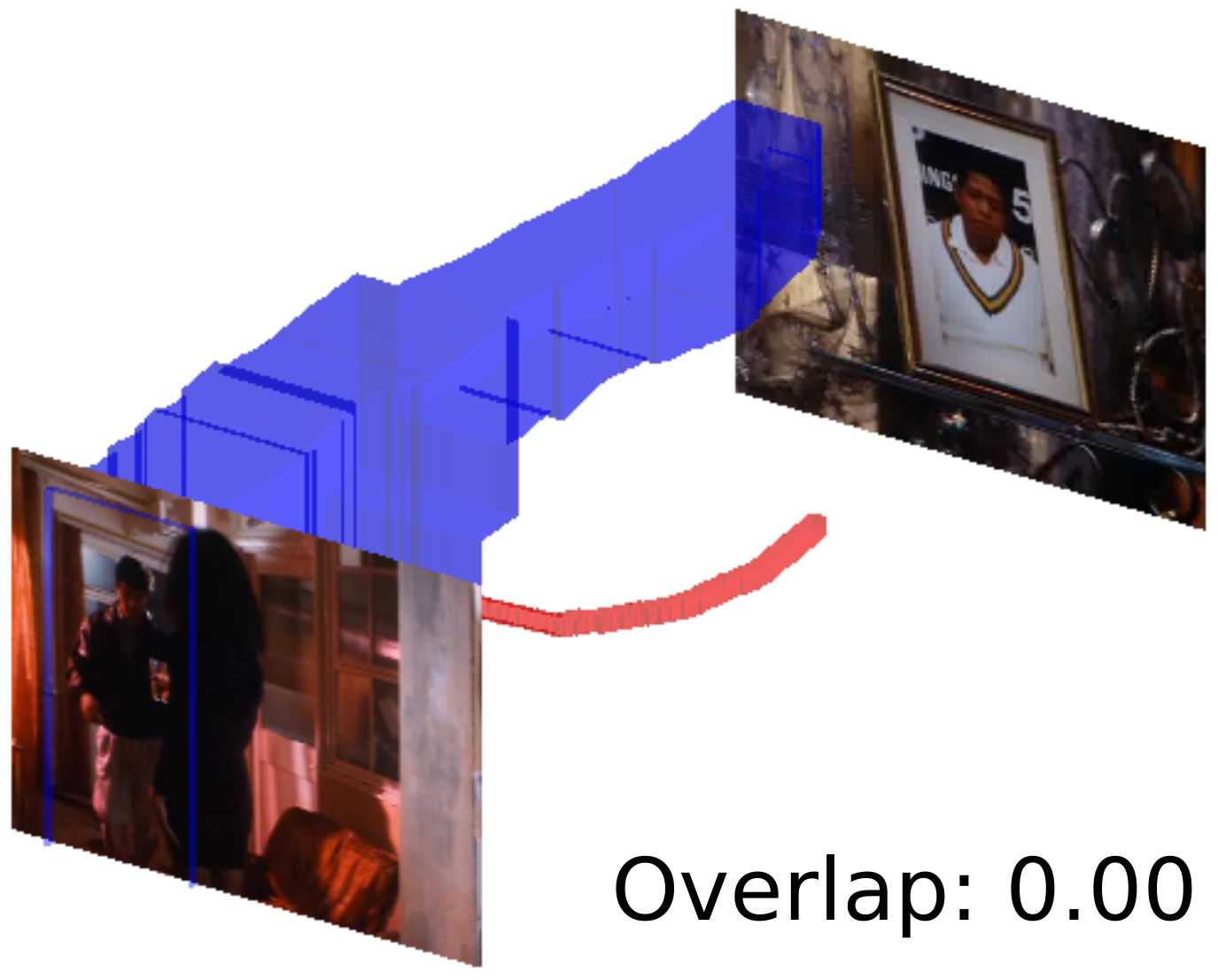}
		\caption{No overlap.}
	\end{subfigure}
	\begin{subfigure}{0.32\linewidth}
		\centering
		\includegraphics[width=\textwidth]{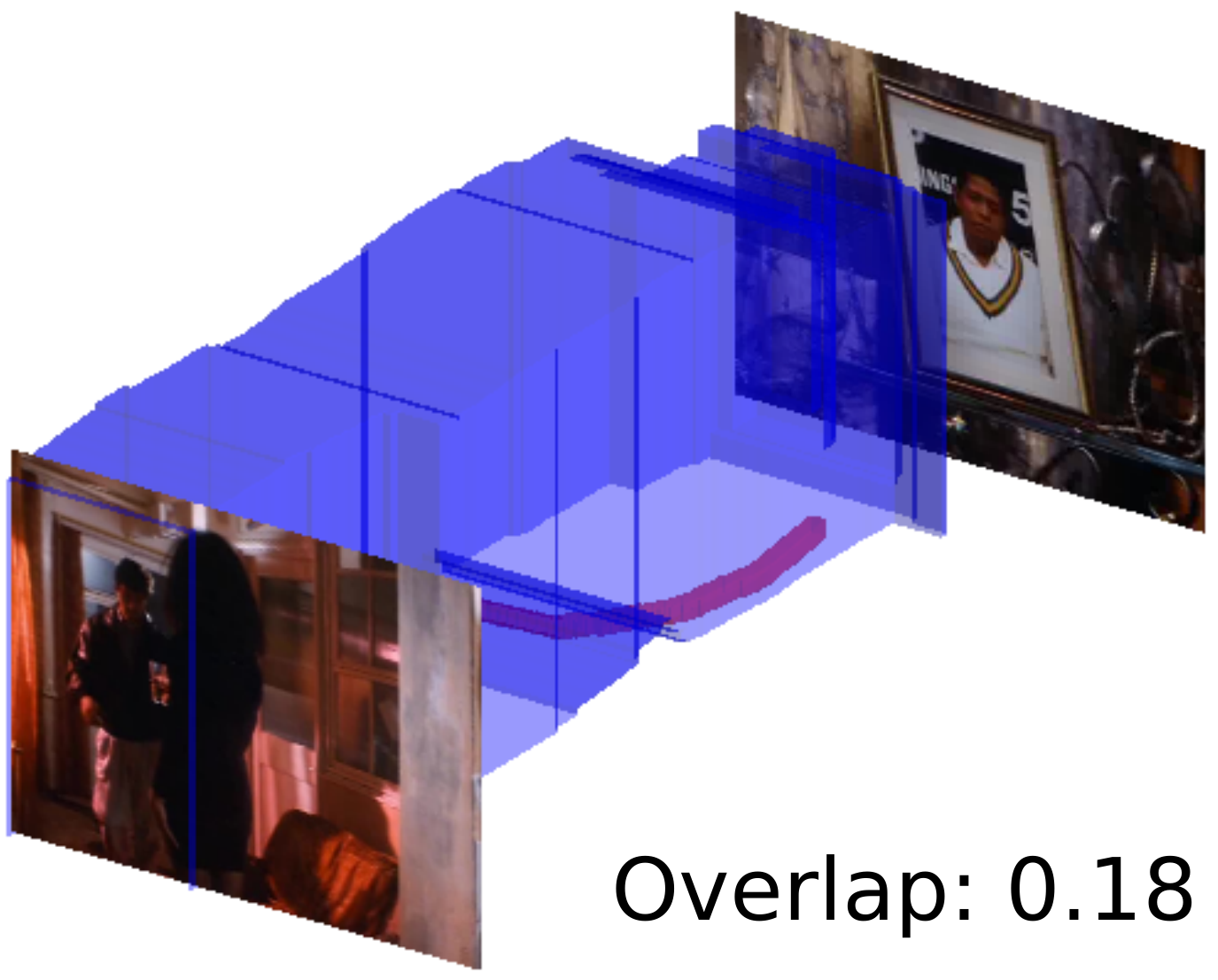}
		\caption{Small overlap.}
	\end{subfigure}
	\begin{subfigure}{0.32\linewidth}
		\centering
		\includegraphics[width=\textwidth]{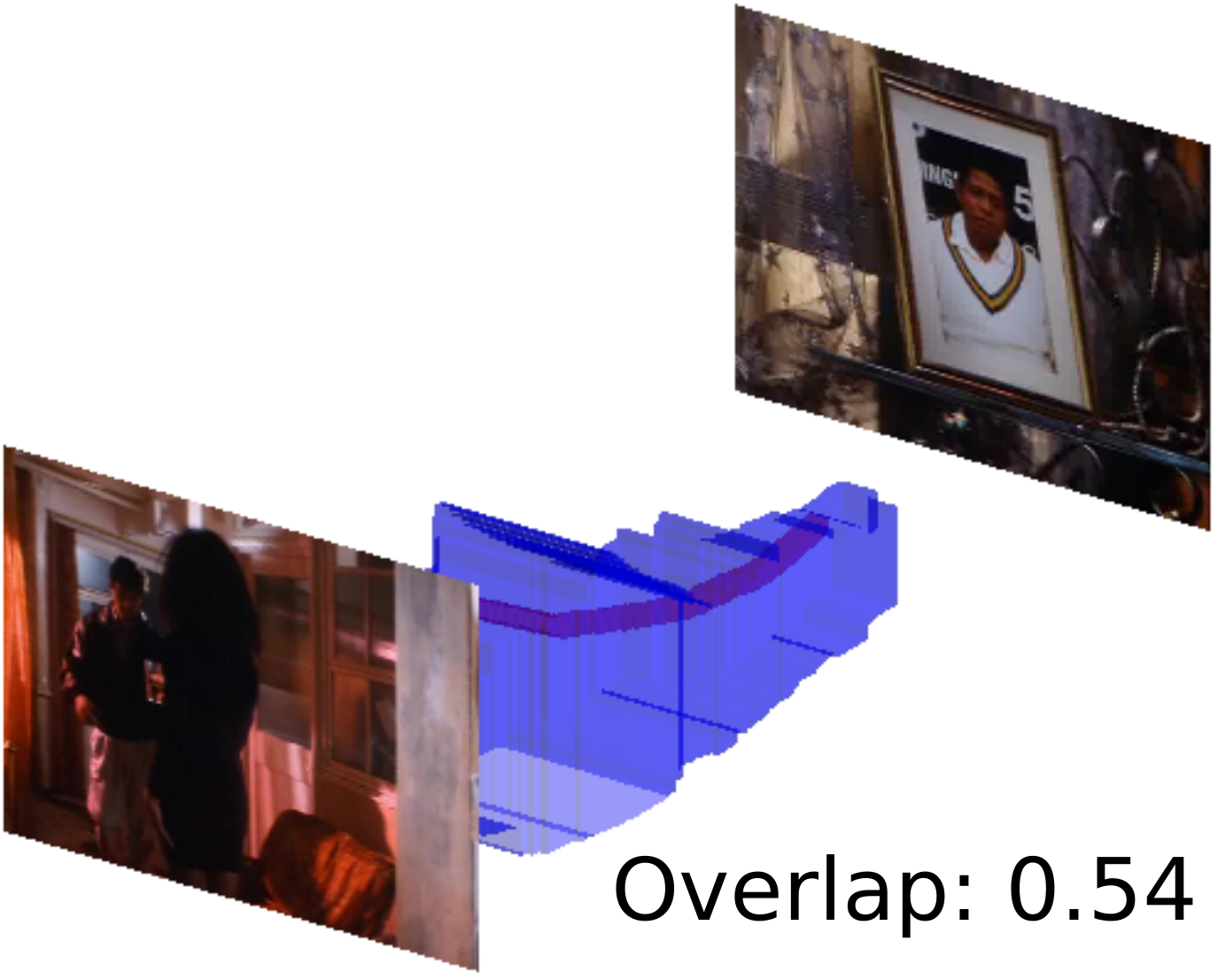}
		\caption{High overlap.}
	\end{subfigure}
	\caption{Illustration of overlap between proposals and points.}
	\label{fig:overlap-example}
\end{figure}

The second term $S(\cdot)$ serves as a form of regularization on the overall size of a proposal.
The regularization aims to alleviate the bias of the first term towards large proposals, since large proposals are more likely to contain points and the box centers of large proposals are by default closer to the center of the video frames.
Since actions are more likely to be in the center of videos~\citep{tseng2009quantifying}, the first term $M(\cdot)$ tends to be biased to large proposals.
The size regularization term $S(\cdot)$ addresses this bias by penalizing proposals with large bounding boxes $|BB_i| \in A$, compared to the size of a video frame $|F_i| \in V$:
\begin{equation}
S(A, V)  =  \bigg( \frac{ \sum_{i=f}^m |BB_i| }{\sum_{i=1}^{|V|} |F_i|} \bigg) ^{2},
\end{equation}
where $|b| = (b(xmax) - b(xmin)) \cdot (b(ymax) - b(ymin))$ denotes the size of box $b$.
Using the center-bias term $M(\cdot)$ regularized by $S(\cdot)$, our overlap measure $O(\cdot)$ is defined as
\begin{equation}
O(A, C, V)  =  M(A, C) - S(A, V).
\label{eq:overlappoint}
\end{equation}
Recall that $A$ are the proposals, $C$ captures the point-supervision and $V$ the video.
Overlap measure $O(\cdot)$ provides an estimation of the quality of the proposals during training and we use the measure in an iterative proposal mining algorithm over all training videos in search for the best proposals. In Figure~\ref{fig:overlap-example}, we provide three visual examples of spatio-temporal proposals ranked based on our overlap measure.

\subsection{Mining for proposals with points}
To mine spatio-temporal proposals, we are given a training set of videos $\{A_{i}, \textbf{x}_{i}, y_{i}, C_i\}_{i=1}^{N}$, where the collection of tubes for the $|A_{i}|$ proposals is denoted as $A_{i} = \{ \{ BB_{j} \}_{j=f}^{m} \}^{|A_{i}|}$. Variable $\textbf{x}_{i} \in \mathbb{R}^{|A_{i}| \times D}$ is the $D$ dimensional feature representation of each proposal in video $i$. Annotations consist of the action class label $y_i$ and the points $C_i$.

The proposal mining combines the use of the overlap measure $O(\cdot)$ of Equation~\ref{eq:overlappoint} with a Multiple Instance Learning optimization.
The optimization aims to train a classification model that can separate good and bad proposals for a given action.
We start from a standard MIL-SVM~\citep{andrews2002support,cinbis2017weakly} and adapt it's objective to include a mining score $P(\cdot)$ of each proposal, which incorporates our function $O(\cdot)$ as:
\begin{equation}
\begin{split}
 & \min_{\mathbf{w},b,\xi} \frac{1}{2} ||\mathbf{w}||^{2} + \lambda \sum_{i} \xi_{i}, \quad \quad \text{s.t.}\\
& \forall_{i} : y_{i} \cdot ( \mathbf{w} \cdot \argmax_{\mathbf{z} \in \mathbf{x}_{i}} P(\mathbf{z} | \mathbf{w}, b, A_{i}^{(\mathbf{z})}, C_{i}, V_i) + b) \geq 1 - \xi_{i},\\
& \forall_{i} : \xi_i \geq 0,
\end{split}
\label{eq:milsvm}
\end{equation}
where $(\mathbf{w},b)$ denote the classifier parameters, $\xi_{i}$ denotes the slack variable and $\lambda$ denotes the regularization parameter.
Variable $\mathbf{z} \in \mathbf{x}_i$ denotes the representation of a single proposal in the set of all proposals $\mathbf{x}_i$ for training video $i$.
Variable $A_{i}^{(\mathbf{z})}$ denotes the tube corresponding to proposal representation $\mathbf{z}$.
The proposal with the highest mining score per video is used to train the classifier.

Different from standard MIL-SVM, the proposals are not only conditioned on the classifier parameters, but also on the overlap scores from the point annotations.
In other words, the standard maximum likelihood optimization of MIL is adapted to include point overlap scores that serve as a prior on the individual proposals.
The objective of Equation~\ref{eq:milsvm} is non-convex due to the joint minimization over the classifier parameters $(\mathbf{w}, b)$ and the maximization over the mined proposals $P(\cdot)$.
Therefore, we perform iterative block coordinate descent by alternating between clamping one and optimizing the other.
Given a fixed selection of proposals, the optimization becomes a standard SVM optimization over the features of selected proposals~\citep{cortes1995support}.
For fixed model parameters, the maximization over the proposals is determined by scoring proposals as:
%
\begin{equation}
\begin{split}
 & P(\mathbf{z} | \mathbf{w}, b, A_{i}^{(\mathbf{z})}, C_{i}, V_i) \propto & \left( <\!\!\mathbf{w}, \mathbf{z}\!\!> + b \right) + \\
 & & O(A_{i}^{(\mathbf{z})}, C_{i}, V_i).\\
\end{split}
\label{eq:mil-e}
\end{equation}
In Equation 5, the score of a proposal is the sum of two components, namely the score of the current model and the overlap with the point annotations in the corresponding training video.
The mining and classifier optimizations are alternated for a fixed amount of iterations. After the iterative optimization, a final SVM is trained on the best mined proposals. Identical to approaches using box-supervision, our model selects the best proposals from test videos, without requiring any box annotations during training.

\begin{figure*}[tbp]
\begin{subfigure}{\textwidth}
\centering
\includegraphics[width=0.23\textwidth]{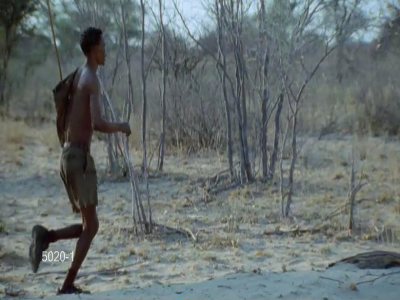}
\hspace{1cm}
\includegraphics[width=0.23\textwidth]{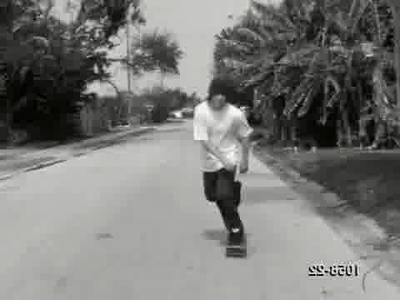}
\hspace{1cm}
\includegraphics[width=0.23\textwidth]{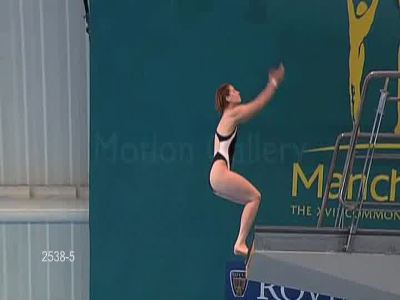}
\caption{Frame.}
\end{subfigure}
\begin{subfigure}{\textwidth}
\centering
\includegraphics[width=0.23\textwidth]{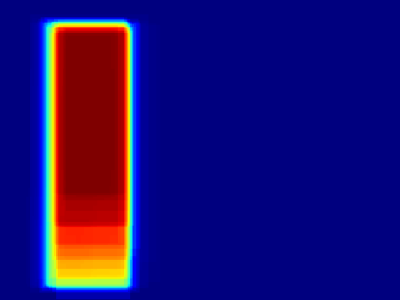}
\hspace{1cm}
\includegraphics[width=0.23\textwidth]{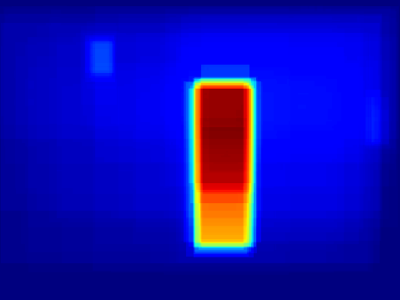}
\hspace{1cm}
\includegraphics[width=0.23\textwidth]{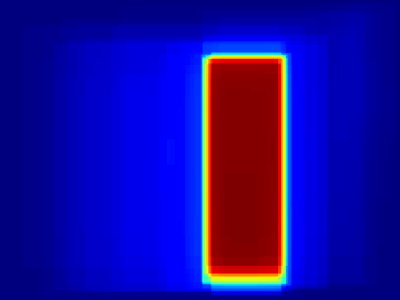}
\caption{Person detection.}
\end{subfigure}
\begin{subfigure}{\textwidth}
\centering
\includegraphics[width=0.23\textwidth]{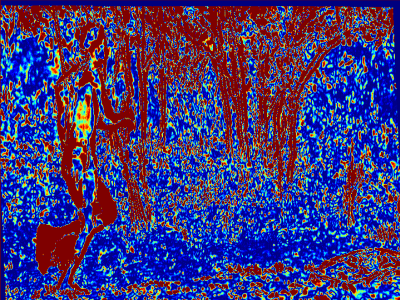}
\hspace{1cm}
\includegraphics[width=0.23\textwidth]{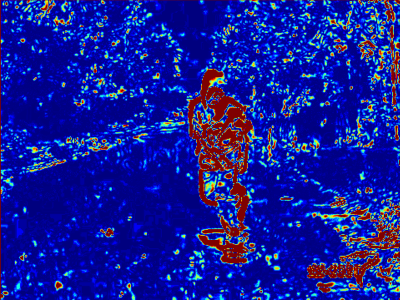}
\hspace{1cm}
\includegraphics[width=0.23\textwidth]{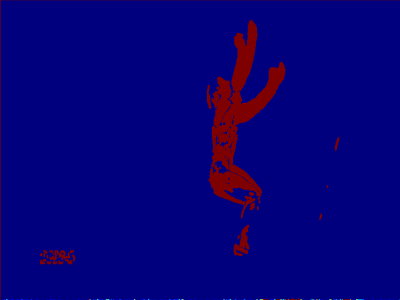}
\caption{Independent motion.}
\end{subfigure}
\begin{subfigure}{\textwidth}
\centering
\includegraphics[width=0.23\textwidth]{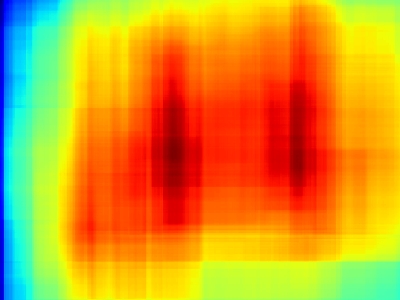}
\hspace{1cm}
\includegraphics[width=0.23\textwidth]{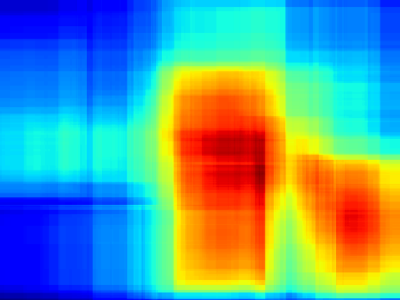}
\hspace{1cm}
\includegraphics[width=0.23\textwidth]{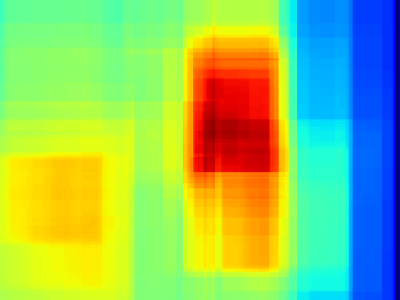}
\caption{Self-supervision.}
\end{subfigure}
\begin{subfigure}{\textwidth}
\centering
\includegraphics[width=0.23\textwidth]{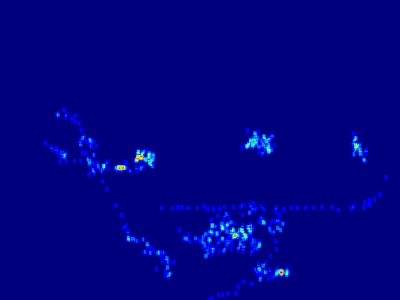}
\hspace{1cm}
\includegraphics[width=0.23\textwidth]{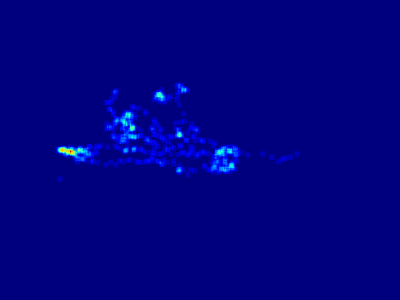}
\hspace{1cm}
\includegraphics[width=0.23\textwidth]{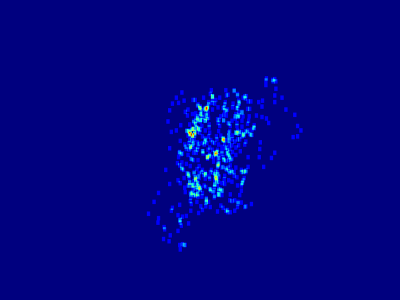}
\caption{Training point statistics.}
\end{subfigure}
\begin{subfigure}{\textwidth}
\centering
\includegraphics[width=0.23\textwidth]{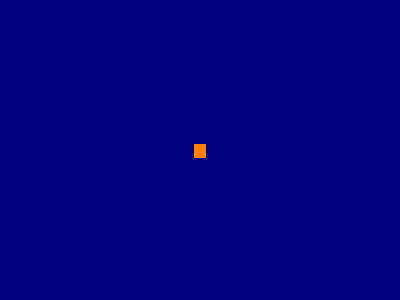}
\hspace{1cm}
\includegraphics[width=0.23\textwidth]{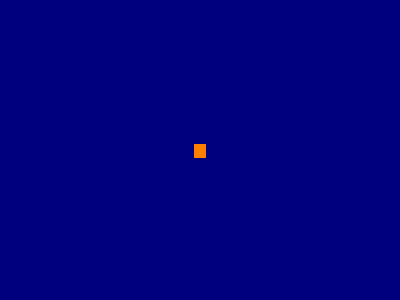}
\hspace{1cm}
\includegraphics[width=0.23\textwidth]{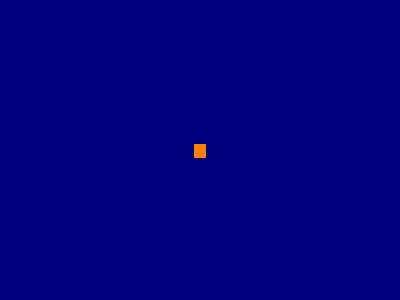}
\caption{Center bias.}
\end{subfigure}
\caption{Pseudo-points for an example frame (a) from three videos showing \emph{running}, \emph{skateboarding}, and \emph{diving}. The pseudo-points derived from (b) person detection, (c) independent motion, and (d) self-supervision focus on the primary action in the video. The pseudo-points derived from (d) training points and (e) center bias provide data-independent prior statistics to steer better proposal selection during inference.}
\label{fig:pseudo-annotations}
\end{figure*}

\section{Pseudo-pointing for inference} \label{sec:method2}
Inference is typically achieved through a maximum likelihood over all proposals in a test video.
However, relying on a maximum likelihood estimate of the model is rather limited, as it only relies on the features of the proposals.
We show that visual cues within the test videos help to guide the selection proposals during inference, similar to how point annotations provide guidance during training. We dub these automatic cues pseudo-points and investigate five of them.
The pseudo-points rely on training point annotations, self-supervision, person detection, independent motion, and center bias. We show how to exploit and combine these pseudo-points to improve the action localization during inference. Lastly, we also provide two forms of regularization to further boost the localization results.

\subsection{The pseudo-points}
In Figure~\ref{fig:pseudo-annotations}, we provide a visual overview of the visual cues for multiple video frames.
Next, we outline each pseudo-point individually.

\subsubsection{Training point statistics}
The first pseudo-point focuses on the point annotations provided during training.
Intuitively, actions do not occur at random locations in video frames.
Recall that we are given $N$ training videos, where $y_i, C_i$ denote respectively the video label and point annotation of training video $i$.
We exploit this observation by making a pseudo-point for an action class $\mathcal{Y}$ as follows:
%
%
\begin{equation}
p_{\text{points}}(F, \mathcal{Y}) = \frac{1}{\sum_{i=1}^{N} [\![ y_i = \mathcal{Y} ]\!]} \sum_{i=1}^{N} [\![ y_i = \mathcal{Y} ]\!] \cdot C_i.
\end{equation}
The above Equation states that for an action $\mathcal{Y}$, the pseudo-point in a test video is determined as the average point annotation location given the training videos of the same action.
The reasoning behind this pseudo-point is that specific actions tend to re-occur in similar locations across videos.
Note that the pseudo-point is independent of the frame $F$ itself and only depends on the training point statistics.

\subsubsection{Self-supervision from proposals}
The second pseudo-point we investigate does not require external information; the pseudo-point relies on the spatio-temporal proposals themselves.
The main idea behind this pseudo-point is that the distribution over all proposals in a single frame provides information about its action uncertainty.
It relies on the following assumption: the more the proposals are on the same spatial location, the higher the likelihood that the action occurs in that location.
The pseudo-point can be seen as a form of self-supervision~\citep{doersch2015unsupervised,fernando2017self}, since it provides an automatic annotation from proposals to guide the selection of the very same proposals.

%
More formally, for test video $t$, let $A_t$ denote the spatio-temporal proposals.
Furthermore, let $F$ denote a video frame and let $C_{A_t}^{\star}(u, v, F)$ denote the number of proposals that contain pixel $(u,v)$ in $F$.
We place an automatic pseudo-point at the center of mass over these pixel counts:
\begin{equation}
p_{\text{self}}(F) = \frac{1}{\sum_{u,v} C_{A_t}^{\star}(x, y, F)} \sum_{(u,v) \in F} C_{A_t}^{\star}(u,v, F) \cdot (u,v).
\label{eq:pss}
\end{equation}
The function of Equation~\ref{eq:pss} outputs a 2D coordinate in frame $F$, representing the center of mass over all pixels in $F$, with the mass of each pixel $(u, v)$ given by $C^{\star}_{A_t}(u,v, F)$. The 2D output coordinate will serve as the pseudo-point in frame $F$.

\subsubsection{Person detection}
The third pseudo-point follows earlier work on action localization by incorporating knowledge about person detections~\citep{siva2011weakly,yuCVPR2015fap}.
Actions are typically person-oriented, so the presence or absence of a person in a proposal provides valuable information.
Here, we employ a Faster R-CNN network~\citep{ren2015faster}, pre-trained on MS-COCO~\citep{lin2014microsoft}, and use the person class for the detections.
This results in roughly 50 box detections per frame after non-maximum suppression.
We select the box in each frame with the maximum confidence score as the automatic pseudo-point.

\subsubsection{Independent motion}
The independent motion of a pixel $(u,v)$ in frame $F$ provides information as to where foreground actions are occurring.
More precisely, independent motion states the deviation from the global motion in a frame~\citep{jain2017tubelets}.
Let $C_{I}^{\star}(u, v, F) \in [0,1]$ denote the inverse of the residual in the global motion estimation at pixel $(u,v)$ in frame $F$.
The higher $C_{I}^{\star}(u, v, F)$, the less likely it is the pixel contributes to the global motion.
Akin to the second pseudo-point, we place an automatic pseudo-point at the center of mass, now over the independent motion estimates:
\begin{equation}
p_{\text{imotion}}(F) = \frac{1}{\sum_{u,v} C_{I}^{\star}(u, v, F)} \sum_{u,v} C_{I}^{\star}(u, v, F) \cdot (u, v).
\label{eq:psi}
\end{equation}
Equation~\ref{eq:psi} outputs a 2D coordinate, but now using the independent motion as mass for each pixel in $F$.

\subsubsection{Direct center bias}
Lastly, we again focus on an observation made during training; actions and annotators have a bias towards the center of the video~\citep{tseng2009quantifying}.
We exploit this bias directly in our fifth pseudo-point by simply placing a point on the center of each frame:
\begin{equation}
p_{\text{center}}(F) = (F_{W} / 2, F_{H} / 2),
\end{equation}
where $F_W$ and $F_H$ denote the width and height of frame $F$ respectively.

\begin{figure*}[t]
	\centering
	\includegraphics[width=0.24\textwidth]{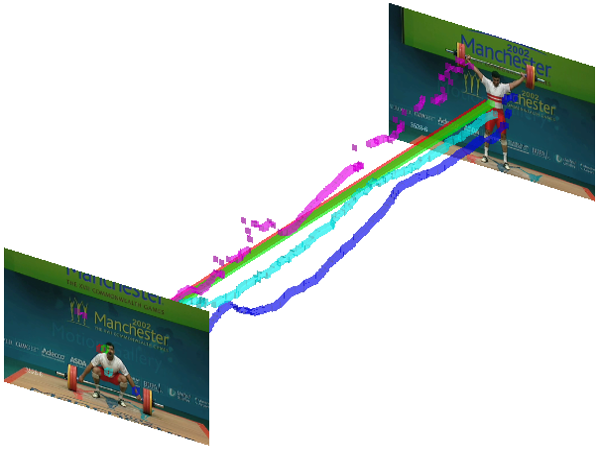}
	\includegraphics[width=0.24\textwidth]{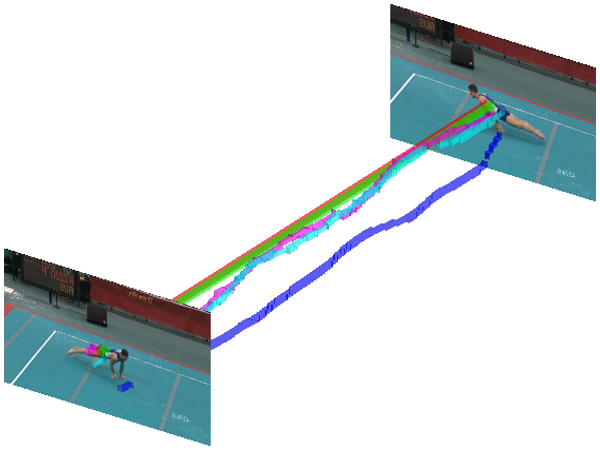}
	\includegraphics[width=0.24\textwidth]{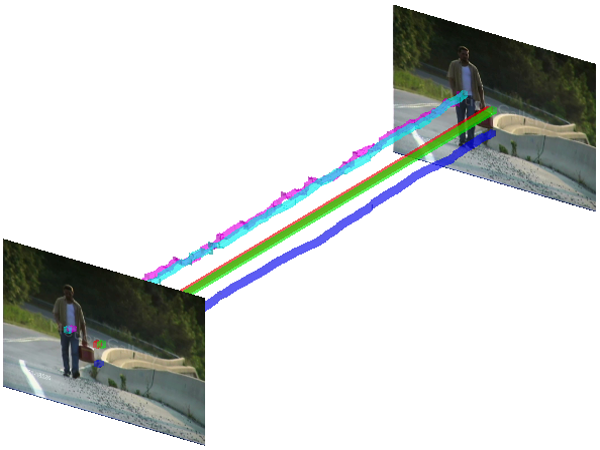}
	\includegraphics[width=0.24\textwidth]{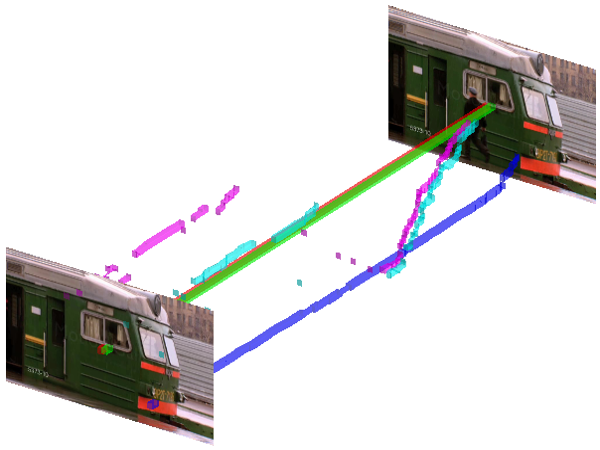}
	\caption{Pseudo-point visualization on four example videos for {\color{red}training points}, {\color{green}center bias}, {\color{blue}self-supervision}, {\color{pink}independent motion}, and {\color{cyan}person detection} (depicted as points for visualization). In general, the pseudo-points are present around the action or even follow the action. When actions are not in the frame however, as shown in the right example, pseudo-points may place automatic annotations in phantom positions.}
	\label{fig:pseudo-evolution}
\end{figure*}

Figure~\ref{fig:pseudo-evolution} provides the spatio-temporal evolution and focus area of the pseudo-points for four example videos.

\subsection{Exploiting pseudo-points}

\subsubsection{Rescoring test proposals}
Given a test video $t$, the standard approach in action localization with spatio-temporal proposal for finding the best proposal is done from a set of proposals $\{A_ti\}_{i=1}^{|A_t|}$, given a model $(\mathbf{w},b)$ is given as:
\begin{equation}
t_{\star} = \argmax_{i=1,..,|A_t|} \left( <\!\!\mathbf{w}, \mathbf{z}_i\!\!> + b \right).
\end{equation}
We exploit the pseudo-points to adjust this likelihood estimate:
\begin{equation}
t_{\star} = \argmax_{i=1,..,|A_t|} \left( <\!\!\mathbf{w}, \mathbf{z}_i\!\!> + b \right) + \lambda_P \cdot O(A_ti, P, t_V),
\label{eq:pa-score}
\end{equation}
where $P$ denotes the pseudo-point of interest and $t_V$ denotes the test video itself.

The above equation is similar in spirit to Equation~\ref{eq:mil-e}, but now automatic cues are employed, rather than manual point annotations. Adjusting the proposal selection using pseudo-points during testing can be seen as a form of regularization. Rather than a single-point maximum likelihood given a trained model, we add continuous restrictions on the proposals based on their match with automatic pseudo-points, which aid the selection towards proposals with a high overlap to the ground truth action location.


\subsubsection{Weighting and selecting pseudo-points}
\label{sec:pa-weights}
Intuitively, not all pseudo-points are equally effective, stating the need for the weights in Equation~\ref{eq:pa-score}.
However, setting proper values for $\lambda_P$ can not be done directly through standard (cross-)validation, as this requires box-supervision.
To overcome both problems, we provide a score function to estimate the quality of each pseudo-point.
This score will be used to both determine which pseudo-point is most favourable to select and directly serve as weighting value in Equation~\ref{eq:pa-score}.

The score function for the person detection pseudo-point (the only pseudo-point that outputs boxes), is identical to the overlap function in Equation~\ref{eq:overlap1}.
This entails that if the center of the top person detection in each frame of a training video matches with the point annotations of the same video, a high score is achieved.
We compute the average match over all training videos as the weight ($\lambda_P$) for person detection. For the other pseudo-points, we are only given points. In these cases, we use the distance to the nearest image border to normalize the distance between the manual point annotation and the automatic pseudo-point annotation. The overall score function is computed in identical fashion as for the person detection.

By matching automatic pseudo-points in training videos with the manual point annotations, we arrive at an automatic quality measure for pseudo-points, which can be used to weight and select pseudo-points.

\subsection{Temporal pseudo-points}

\begin{figure*}[t]
	\centering
	\begin{subfigure}{0.49\textwidth}
		\begin{minipage}{0.32\textwidth}
			\includegraphics[width=\linewidth]{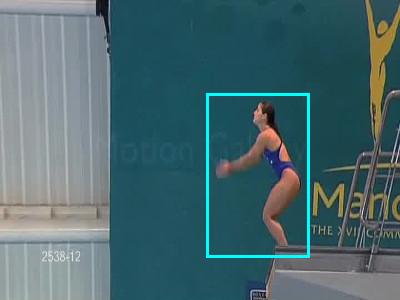} \begin{center} \vspace{-0.2cm} Dive \end{center}
		\end{minipage}
		\begin{minipage}{0.32\textwidth}
			\includegraphics[width=\linewidth]{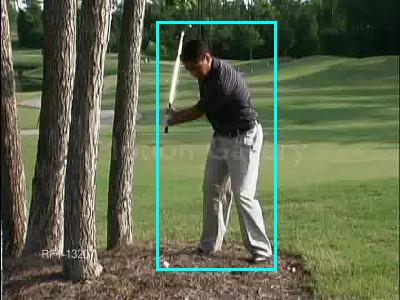} \begin{center} \vspace{-0.2cm} Golf swing \end{center}
		\end{minipage}
		\begin{minipage}{0.32\textwidth}
			\includegraphics[width=\linewidth]{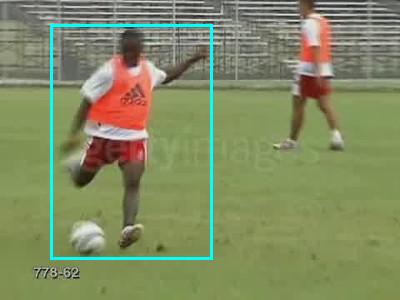} \begin{center} \vspace{-0.2cm} Kick \end{center}
		\end{minipage}
		\\
		\begin{minipage}{0.32\textwidth}
			\includegraphics[width=\linewidth]{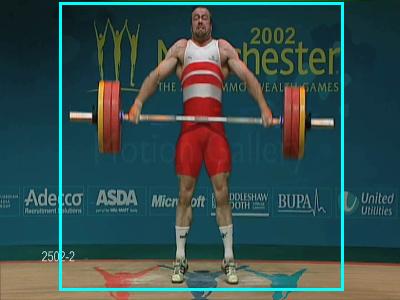} \begin{center} \vspace{-0.2cm} Lift \end{center}
		\end{minipage}
		\begin{minipage}{0.32\textwidth}
			\includegraphics[width=\linewidth]{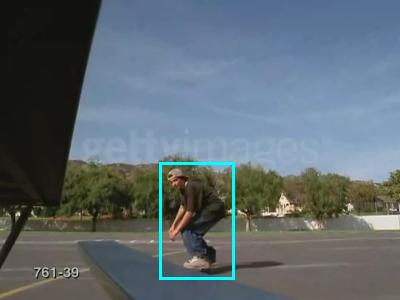} \begin{center} \vspace{-0.2cm} Skateboard \end{center}
		\end{minipage}
		\begin{minipage}{0.32\textwidth}
			\includegraphics[width=\linewidth]{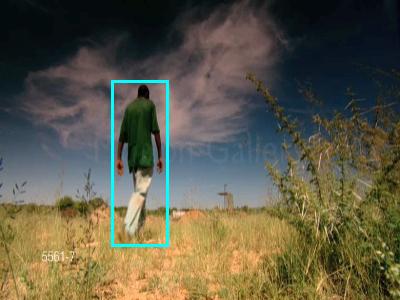} \begin{center} \vspace{-0.2cm} Walk \end{center}
		\end{minipage}
		\caption{UCF Sports.}
		\label{fig:datasets-ucfsports}
	\end{subfigure}
	\begin{subfigure}{0.49\textwidth}
		\begin{minipage}{0.32\textwidth}
			\includegraphics[width=\linewidth]{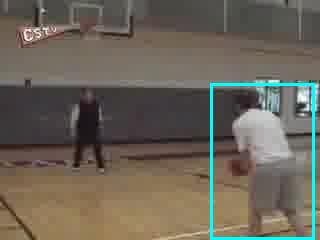} \begin{center} \vspace{-0.2cm} Basketball \end{center}
		\end{minipage}
		\begin{minipage}{0.32\textwidth}
			\includegraphics[width=\linewidth]{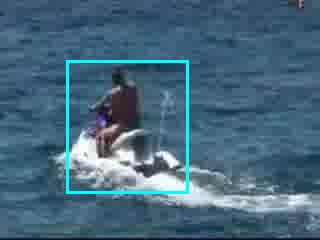} \begin{center} \vspace{-0.2cm} Skijet \end{center}
		\end{minipage}
		\begin{minipage}{0.32\textwidth}
			\includegraphics[width=\linewidth]{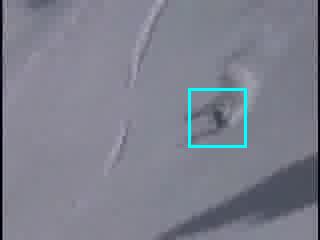} \begin{center} \vspace{-0.2cm} Ski \end{center}
		\end{minipage}
		\\
		\begin{minipage}{0.32\textwidth}
			\includegraphics[width=\linewidth]{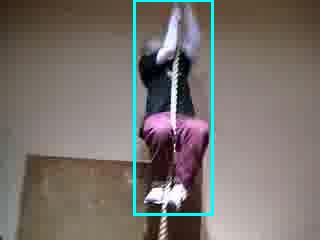} \begin{center} \vspace{-0.2cm} Rope climb \end{center}
		\end{minipage}
		\begin{minipage}{0.32\textwidth}
			\includegraphics[width=\linewidth]{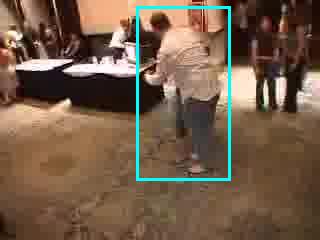} \begin{center} \vspace{-0.2cm} Salsa spin \end{center}
		\end{minipage}
		\begin{minipage}{0.32\textwidth}
			\includegraphics[width=\linewidth]{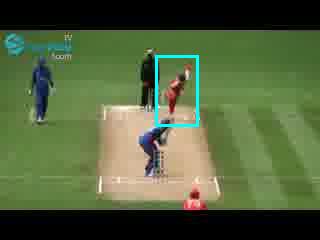} \begin{center} \vspace{-0.2cm} Cricket bowl \end{center}
		\end{minipage}
		\caption{UCF-101.}
		\label{fig:datasets-ucf101}
	\end{subfigure}
	\begin{subfigure}{\textwidth}
		\centering
		\begin{minipage}{0.16\textwidth}
			\includegraphics[width=\linewidth]{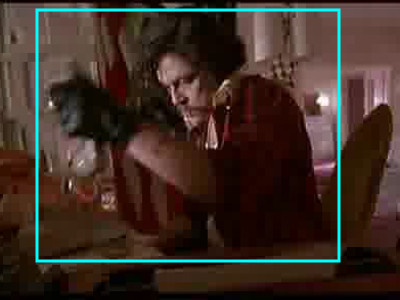} \begin{center} \vspace{-0.2cm} Answer phone \end{center}
		\end{minipage}
		\begin{minipage}{0.16\textwidth}
			\includegraphics[width=\linewidth]{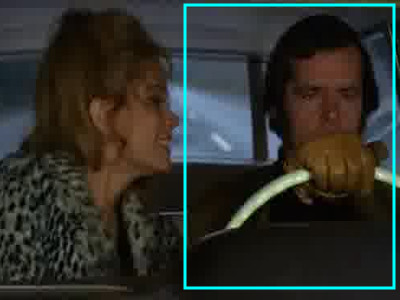} \begin{center} \vspace{-0.2cm} Drive car \end{center}
		\end{minipage}
		\begin{minipage}{0.16\textwidth}
			\includegraphics[width=\linewidth]{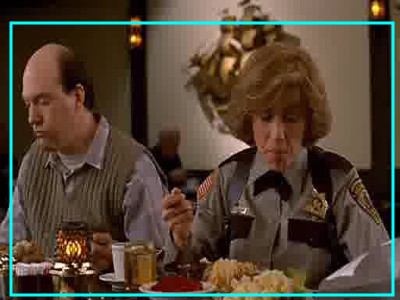} \begin{center} \vspace{-0.2cm} Eat \end{center}
		\end{minipage}
		\begin{minipage}{0.16\textwidth}
			\includegraphics[width=\linewidth]{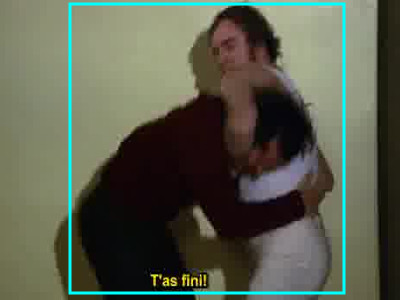} \begin{center} \vspace{-0.2cm} Fight \end{center}
		\end{minipage}
		\begin{minipage}{0.16\textwidth}
			\includegraphics[width=\linewidth]{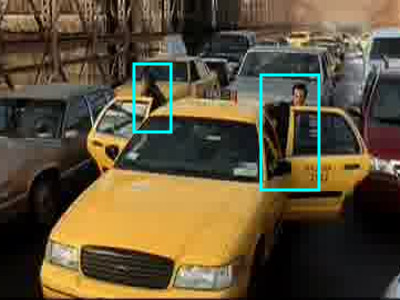} \begin{center} \vspace{-0.2cm} Get our of car \end{center}
		\end{minipage}
		\begin{minipage}{0.16\textwidth}
			\includegraphics[width=\linewidth]{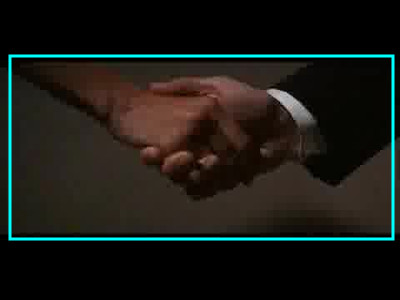} \begin{center} \vspace{-0.2cm} Shake hand \end{center}
		\end{minipage}
		\\
		\begin{minipage}{0.16\textwidth}
			\includegraphics[width=\linewidth]{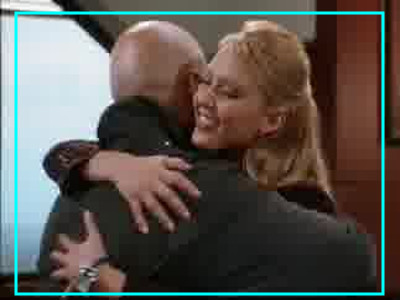} \begin{center} \vspace{-0.2cm} Hug \end{center}
		\end{minipage}
		\begin{minipage}{0.16\textwidth}
			\includegraphics[width=\linewidth]{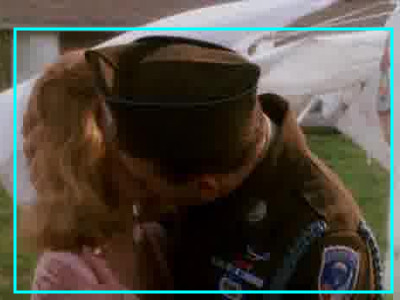} \begin{center} \vspace{-0.2cm} Kiss \end{center}
		\end{minipage}
		\begin{minipage}{0.16\textwidth}
			\includegraphics[width=\linewidth]{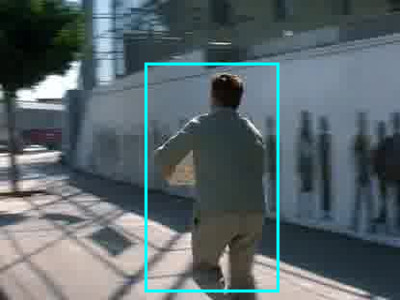} \begin{center} \vspace{-0.2cm} Run \end{center}
		\end{minipage}
		\begin{minipage}{0.16\textwidth}
			\includegraphics[width=\linewidth]{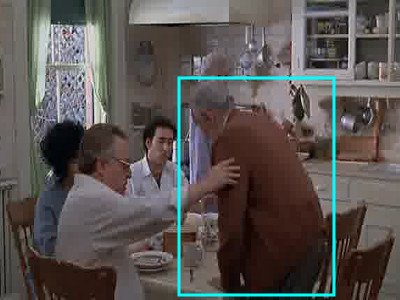} \begin{center} \vspace{-0.2cm} Sit down \end{center}
		\end{minipage}
		\begin{minipage}{0.16\textwidth}
			\includegraphics[width=\linewidth]{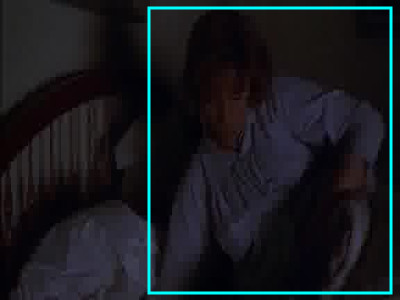} \begin{center} \vspace{-0.2cm} Sit up \end{center}
		\end{minipage}
		\begin{minipage}{0.16\textwidth}
			\includegraphics[width=\linewidth]{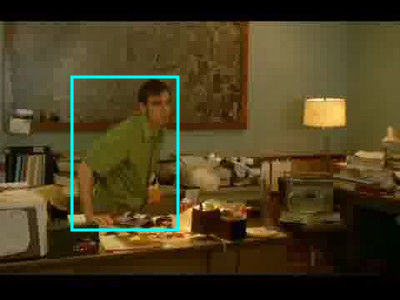} \begin{center} \vspace{-0.2cm} Stand up \end{center}
		\end{minipage}
		\caption{Hollywood2Tubes.}
		\label{fig:datasets-h2t}
	\end{subfigure}
	\caption{Example frames of the UCF Sports, UCF-101, and Hollywood2Tubes dataset with box annotations.
		Different from the UCF Sports and UCF-101 datasets, Hollywood2Tubes provides new challenges for spatio-temporal action localization, due to large occlusions, small inter-class difference, and large action size variation.}
	\label{fig:dataset-examples}
\end{figure*}

Besides knowing where specific actions occur spatially over a complete dataset, the temporal extent of actions is also helpful for proposal selection.
Here, we provide a temporal pseudo-point, again relying on training point statistics.
For action $\mathcal{Y}$, we retrieve its temporal extent by comparing the temporal span of point annotations to the temporal extent of the videos in which the actions occur.
The fraction of the annotation span relative to the video length is computed for each action instance and averaged over a complete dataset.
Let $\mathcal{F}_\mathcal{Y}$ denote the average temporal length of action $\mathcal{Y}$ and let $\mathcal{F}_{t_{jk}}$ denote the temporal length of proposal $k$ in test video $j$.
Then we compute this temporal pseudo-point as:
\begin{equation}
\begin{split}
s^{\mathcal{Y}\star}_{j} = \argmax_{k \in \{1,..,|\mathbf{x}|\}} & \bigg( \left( <\!\!\mathbf{w}_j, \mathbf{x}_k\!\!> + b_j \right) -\\
& \lambda_{T} \cdot \frac{| \mathcal{F}_{\mathcal{Y}} - \mathcal{F}_{t_{jk}} |}{\mathcal{F}_{\mathcal{Y}}} \bigg).\\
\end{split}
\label{eq:temppa}
\end{equation}
In Equation~\ref{eq:temppa}, the match between the temporal pseudo-point of an action and the temporal extent of a proposal also acts as a regularization.
The better the match, the lower the penalty in the likelihood, resulting in a better selection of proposals.

With pseudo-points, we are able to guide the selection of the top proposal per action per video for action localization, akin to how point-supervision is used during training. Having defined the complete pointly-supervised regime for training and inference we are now ready for the experiments.


%
%
\section{Experimental setup} \label{sec:exp1}

\subsection{Datasets}

\textbf{UCF Sports.}
The UCF Sports dataset consists of 150 videos from 10 sport action categories, such as \emph{Skateboarding}, \emph{Horse riding}, and \emph{Walking}~\citep{RodriguezCVPR2008}.
We employ the train/test split suggested in~\citep{lan2011discriminative}.
Example frames from the dataset are shown in Figure~\ref{fig:datasets-ucfsports}.

\textbf{UCF-101.}
The UCF-101 dataset consists of 13,320 videos from 101 action categories, such as \emph{Skiing}, \emph{Basketball dunk}, and \emph{Surfing}~\citep{soomro2012ucf101}.
For a subset of 3,204 videos and 24 categories, spatio-temporal annotations are provided~\citep{soomro2012ucf101}.
We will use this subset in the experiments and use the first train/test split suggested in~\citep{soomro2012ucf101}.
In Figure~\ref{fig:datasets-ucf101}, we show dataset example frames.

\textbf{Hollywood2Tubes.}
The Hollywood2Tubes dataset consists of 1,707 videos from 12 action categories~\citep{mettes2016spot}, such as \emph{getting out of a car}, \emph{sitting down}, and \emph{eating}.
The dataset is derived from the Hollywood2 dataset~\citep{marszalek09}, with point annotations for training and box annotations for evaluation.
Different from current action localization datasets, Hollywood2Tubes is multi-label and actions can be multi-shot, \ie can span over multiple non-continuous shot, adding new challenges for action localization.
We show an example frame with box annotations for each of the 12 actions in the dataset in Figure~\ref{fig:datasets-h2t}.
Annotations are available at \url{http://tinyurl.com/hollywood2tubes}.

\begin{figure*}[t]
\centering
\begin{subfigure}{0.49\textwidth}
\centering
\includegraphics[width=0.49\linewidth]{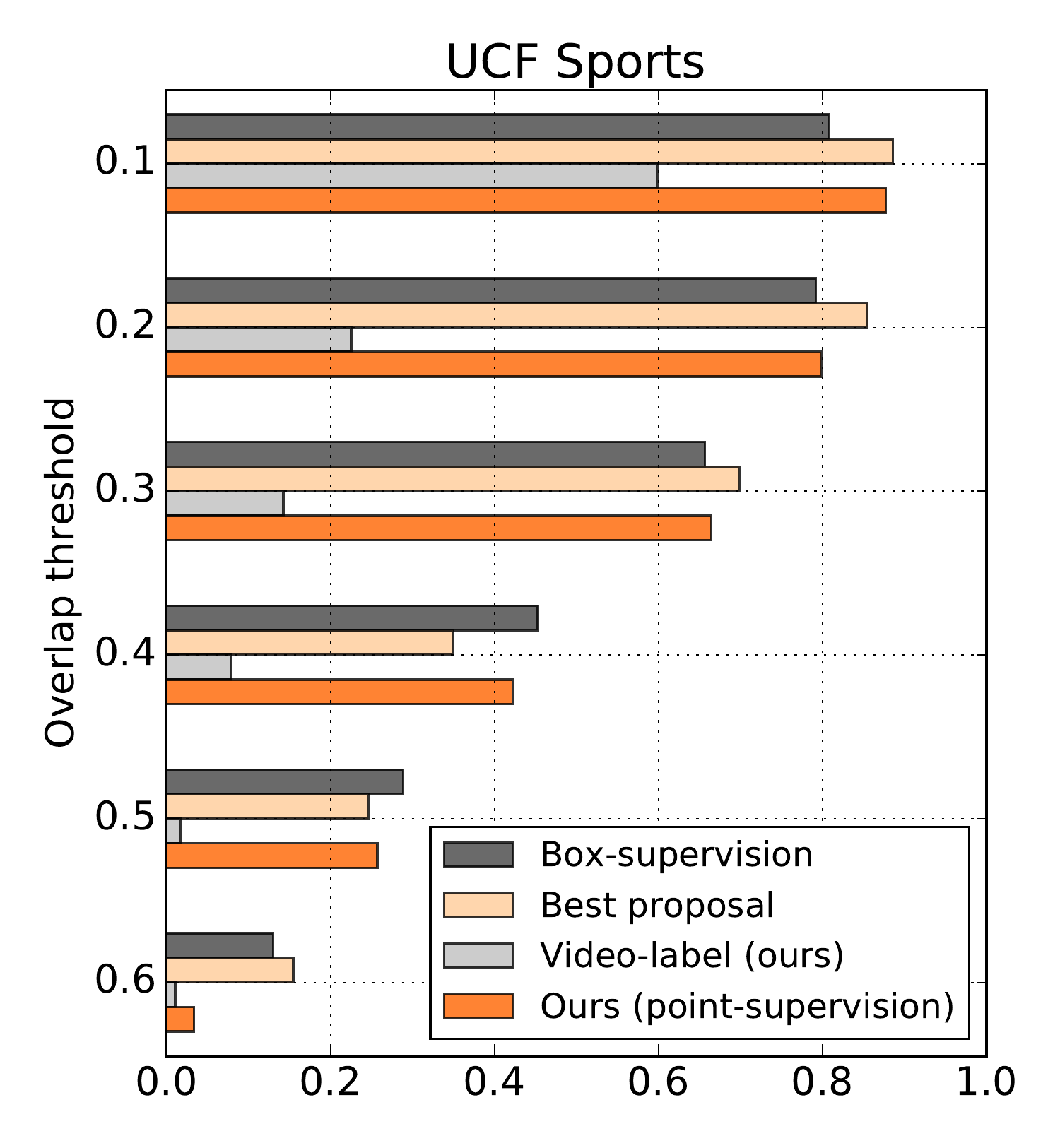}
\includegraphics[width=0.49\linewidth]{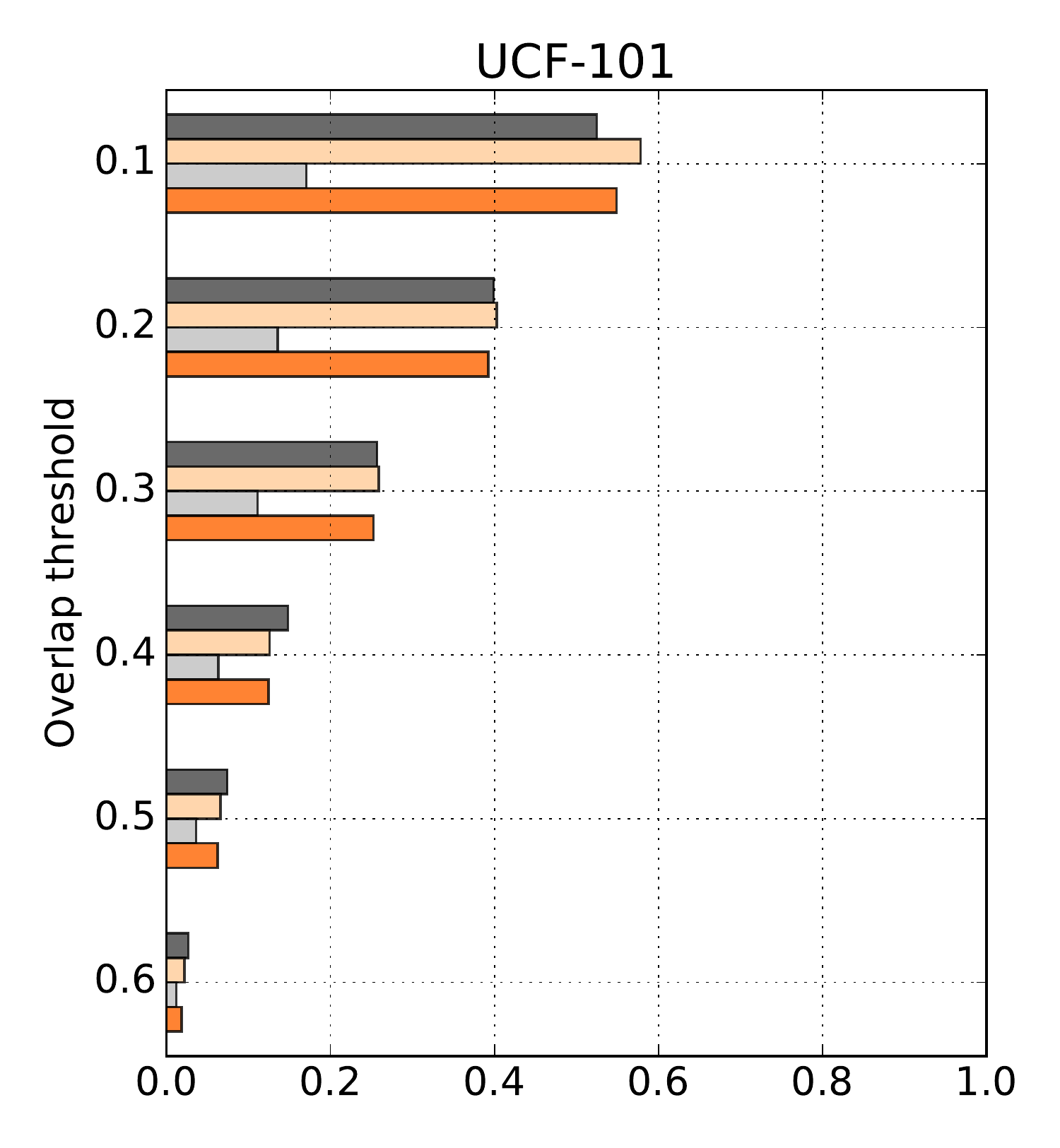}
\caption{mean Average Precision.}
\label{fig:exp1-map}
\end{subfigure}
\begin{subfigure}{0.49\textwidth}
\centering
\includegraphics[width=0.49\linewidth]{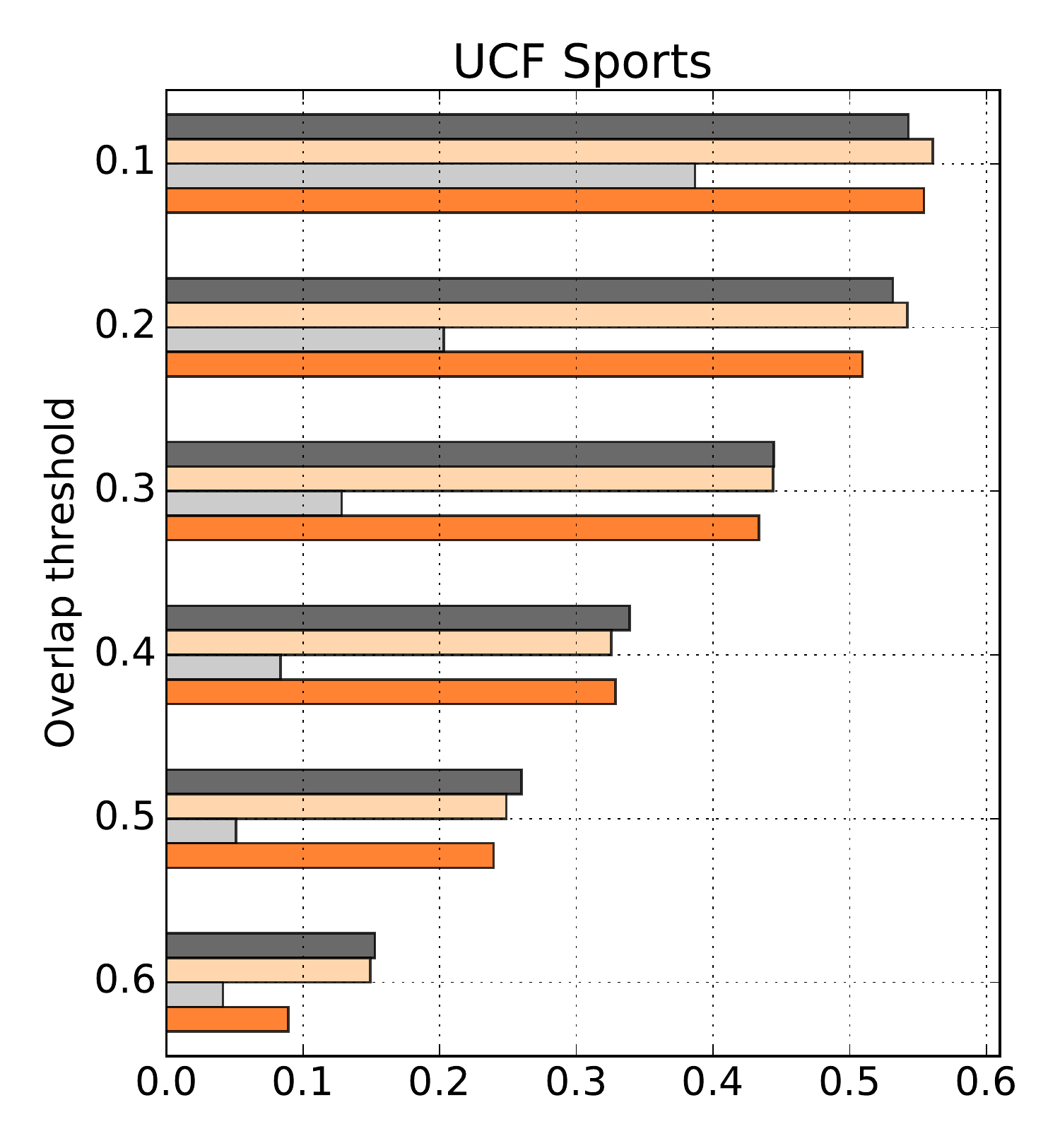}
\includegraphics[width=0.49\linewidth]{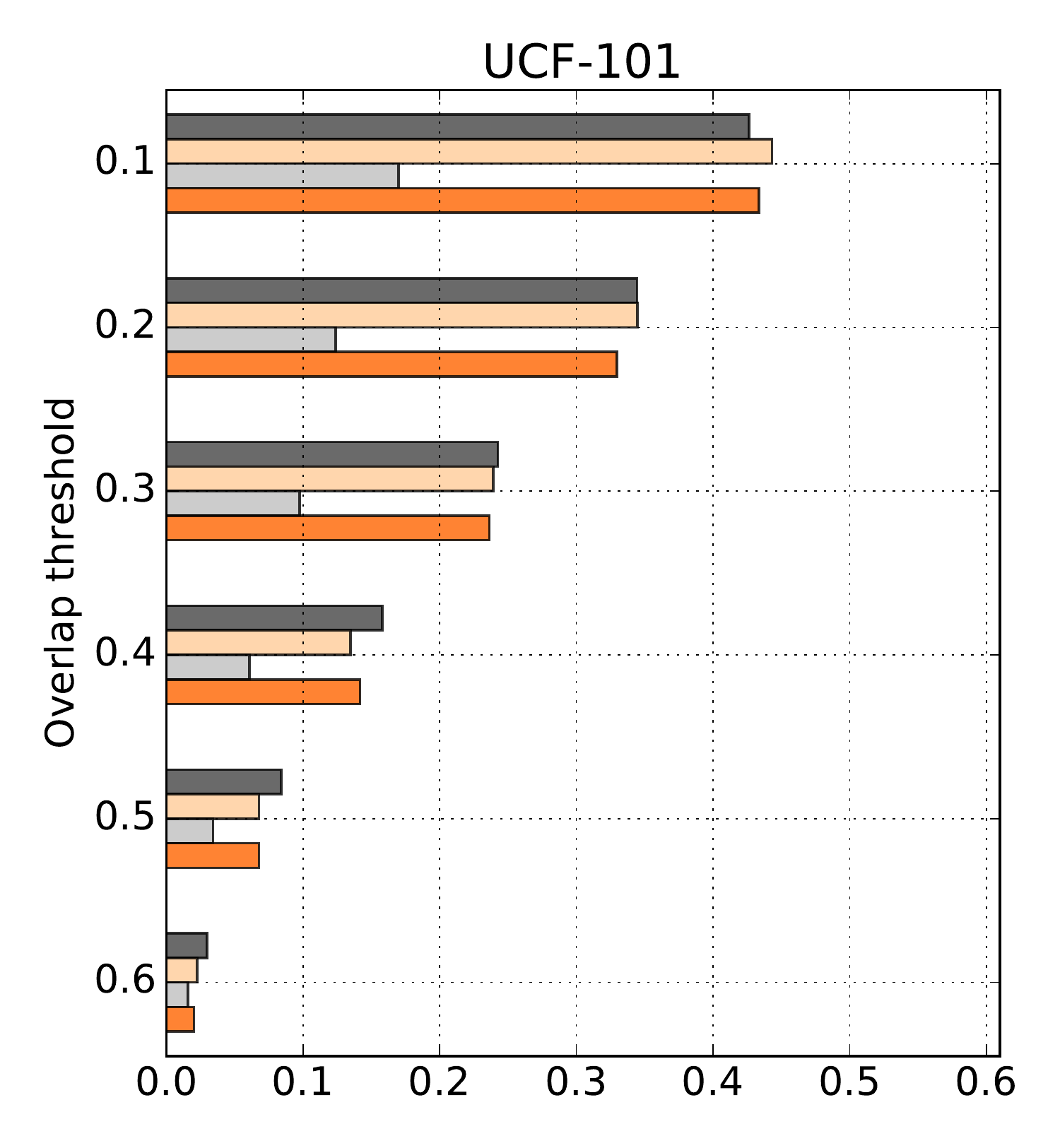}
\caption{AUC.}
\label{fig:exp1-auc}
\end{subfigure}
\caption{Localization results on UCF Sports and UCF-101 using box-supervision, point-supervision, and video-labels only.
Across both datasets and all overlap thresholds, point-supervision is as effective as box-supervision, while they both outperform video-label supervision. We conclude that spatial annotations are vital and that points provide sufficient support for effective localization.}
\label{fig:exp1-res}
\end{figure*}

\subsection{Implementation details}
\textbf{Proposals.}
Our proposal mining algorithm is agnostic to the underlying spatio-temporal proposal algorithm.
Through this work, we employ the unsupervised APT proposals~\citep{gemert2015apt}, since the algorithm provides high action recall, is fast to execute, and the code is publicly available.  For each proposal, we extract Improved Dense Trajectories and compute HOG, HOF, Traj, and MBH features~\citep{wang13}. The combined features are concatenated and aggregated into a fixed-size representation using Fisher Vectors~\citep{sanchez2013image}. We construct a codebook of 128 clusters, resulting in a 54,656-dimensional representation per proposal. The same proposals and representations are also used in \citep{gemert2015apt,mettes2016spot,mettes2017localizing} allowing for a fair comparison.


\textbf{Training.}
The proposal mining is performed for 5 iterations; more iterations have little effect on performance.
Following the suggestions of \cite{cinbis2017weakly}, the training videos are randomly split into 3 splits to train and select the proposals.
For training a classification for one action, 100 proposals of each video are randomly sampled from the other actions as negatives.
The regularization parameter $\lambda$ in the max-margin optimization is fixed to 10 throughout the experiments.

\textbf{Evaluation.}
For an action, we select the top scoring proposal for each test video given the trained model.
To evaluate the action localization performance, we compute the Intersection-over-Union (IoU) between proposal $p_{1}$ and the box annotations of the corresponding test example $p_{2}$ as: $\text{iou}(p_{1}, p_{2}) = \frac{1}{| \Gamma |} \sum_{f \in \Gamma} IoU_{p_{1}, p_{2}}(f)$, where $\Gamma$ is the set of frames where at least one of $(p_{1}, p_{2})$ is present.
The function $IoU$ states the box overlap within a specified frame.
%
For IoU threshold $\tau$, a top selected proposal is deemed a positive detection if $\text{iou}(p_{1}, p_{2}) \geq \tau$.
After aggregating the top tubes from all videos, we compute either the Average Precision score or AUC using the proposal scores and positive/negative detection labels.

%
%
\section{Results} \label{sec:exp2}

%
%
\subsection{Action localization with point-supervision}
\textbf{Setup.}
In the first experiment, we evaluate our main notion of localizing actions using point-supervision.
We perform this evaluation on UCF Sports and UCF-101. We compare our approach to the following three baselines:
\begin{itemize}
\item \texttt{box-supervision}: This baseline follows the training protocol of~\cite{gemert2015apt}, where for each action, a classifier is trained using the features from ground truth boxes. Additionally, spatio-temporal proposals with an overlap higher than 0.6 and lower than 0.1 are added as positives and negatives, respectively.
\item \texttt{best proposal}: This baseline trains an action localizer using the spatio-temporal proposal with the highest overlap to the ground truth box tube.
\item \texttt{video label (ours)}: This baseline employs MIL optimization with a uniform prior, \ie only video labels are used as annotations. This baseline is inspired by~\citep{cinbis2017weakly}, but performed on action proposals in videos instead of object proposals in images.
\end{itemize}
Unless stated otherwise, we employ the centers of the original box annotations on UCF Sports and UCF-101 as the point annotations throughout our experiments.
\\\\
\textbf{Results.}
The results on UCF Sports and UCF-101 are shown in Figure~\ref{fig:exp1-map} (mean Average Precision) and Figure~\ref{fig:exp1-auc} (AUC).
We first observe that traditional box-supervision yield identical results to using the best possible spatio-temporal proposal.
This result validates our starting hypothesis that spatio-temporal proposals provide viable training examples.
Second, we observe that across both datasets and all overlap thresholds, point-supervision performs similar to both the box-supervision and best proposal approaches. This result highlights the effectiveness of point-supervision for action localization.
With pointly-supervised action localization we no longer require expensive box annotations. As results using video-labels only are limited compared to points, we conclude that points provide vital information about the spatial location of actions.
\\\\
\begin{figure}[t]
\centering
\begin{subfigure}{0.49\textwidth}
\centering
\includegraphics[width=\linewidth]{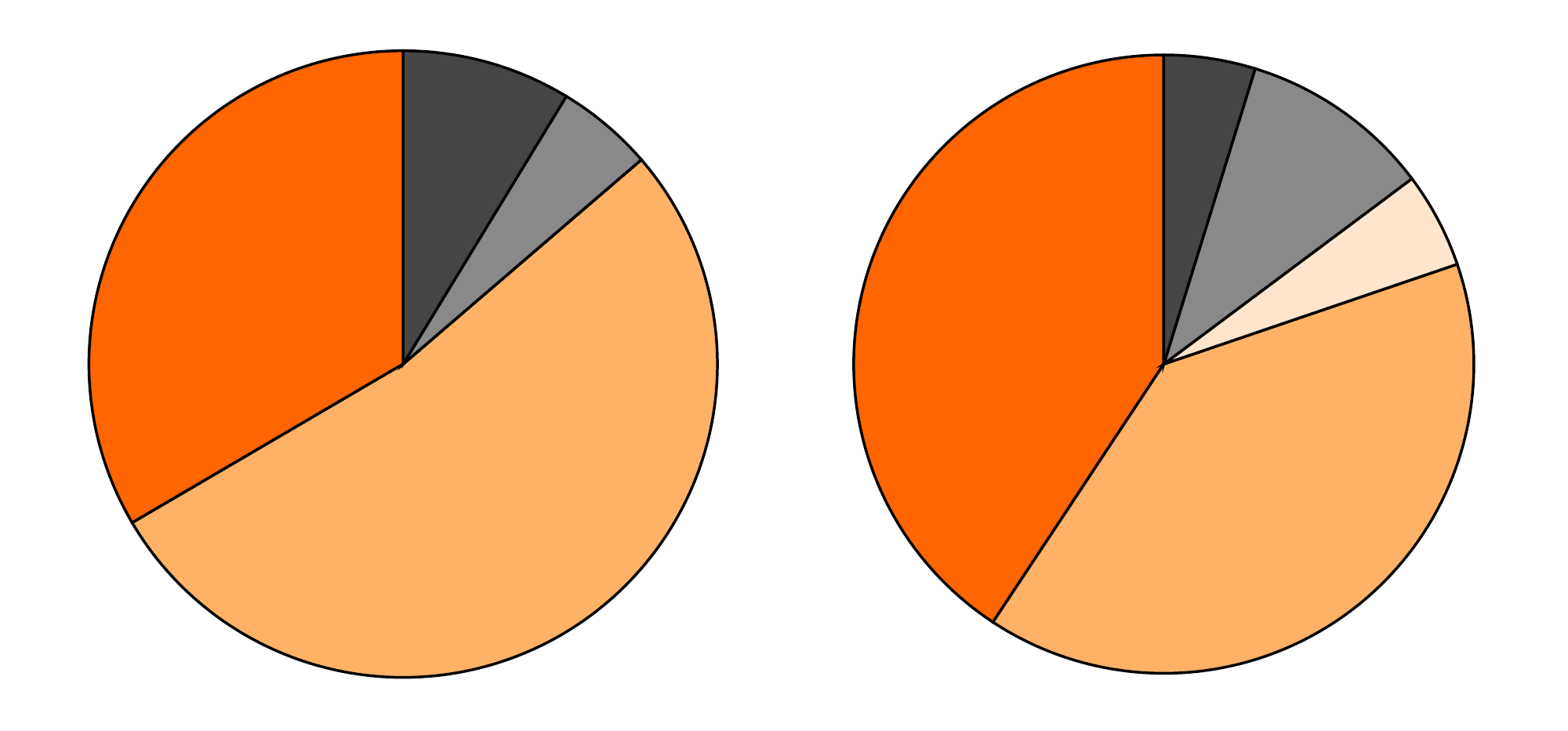}
\caption{UCF Sports.}
\end{subfigure}
\begin{subfigure}{0.49\textwidth}
\centering
\includegraphics[width=\linewidth]{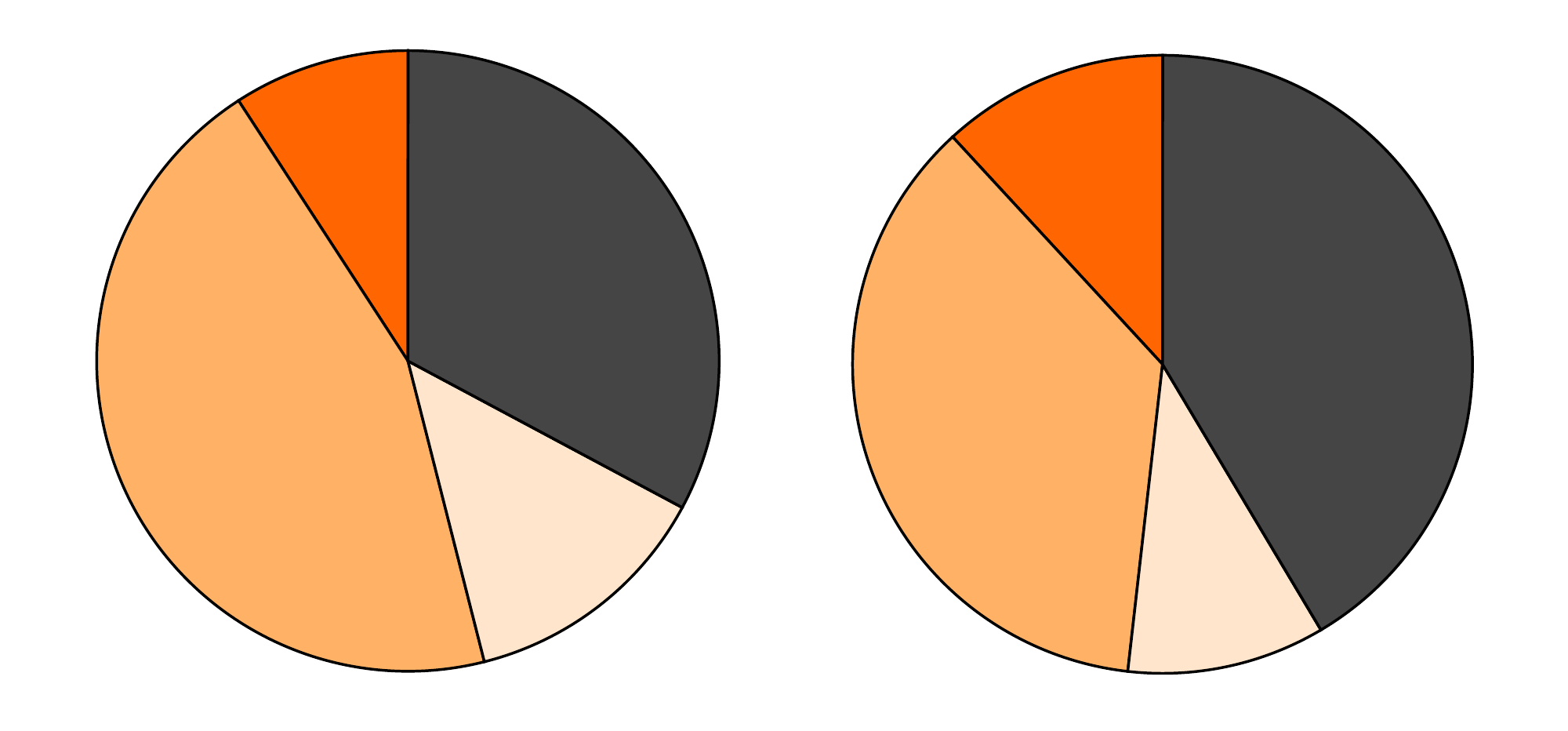}
\includegraphics[width=\linewidth]{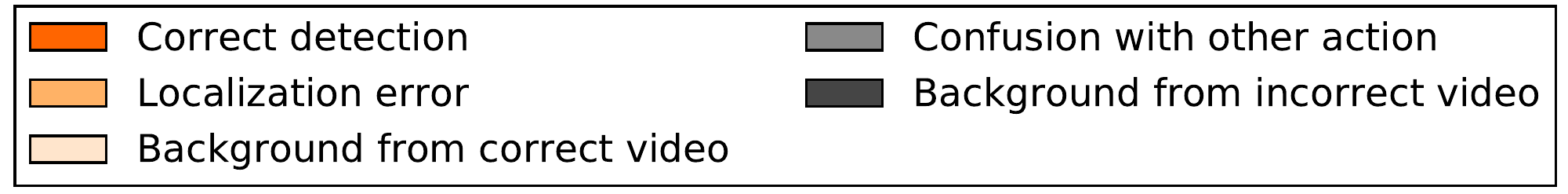}
\caption{UCF-101.}
\end{subfigure}
\caption{Action localization error diagnosis on UCF Sports and UCF-101 when using point-supervision (left) and box-supervision (right).
On both datasets, we observe that averaged over all actions, approaches using point- and box-supervision yield similar error type distributions, explaining their similar localization performance.}
\label{fig:exp1-error-avg}
\end{figure}
\begin{figure*}[t]
\begin{subfigure}{\textwidth}
\centering
\includegraphics[width=0.3\textwidth]{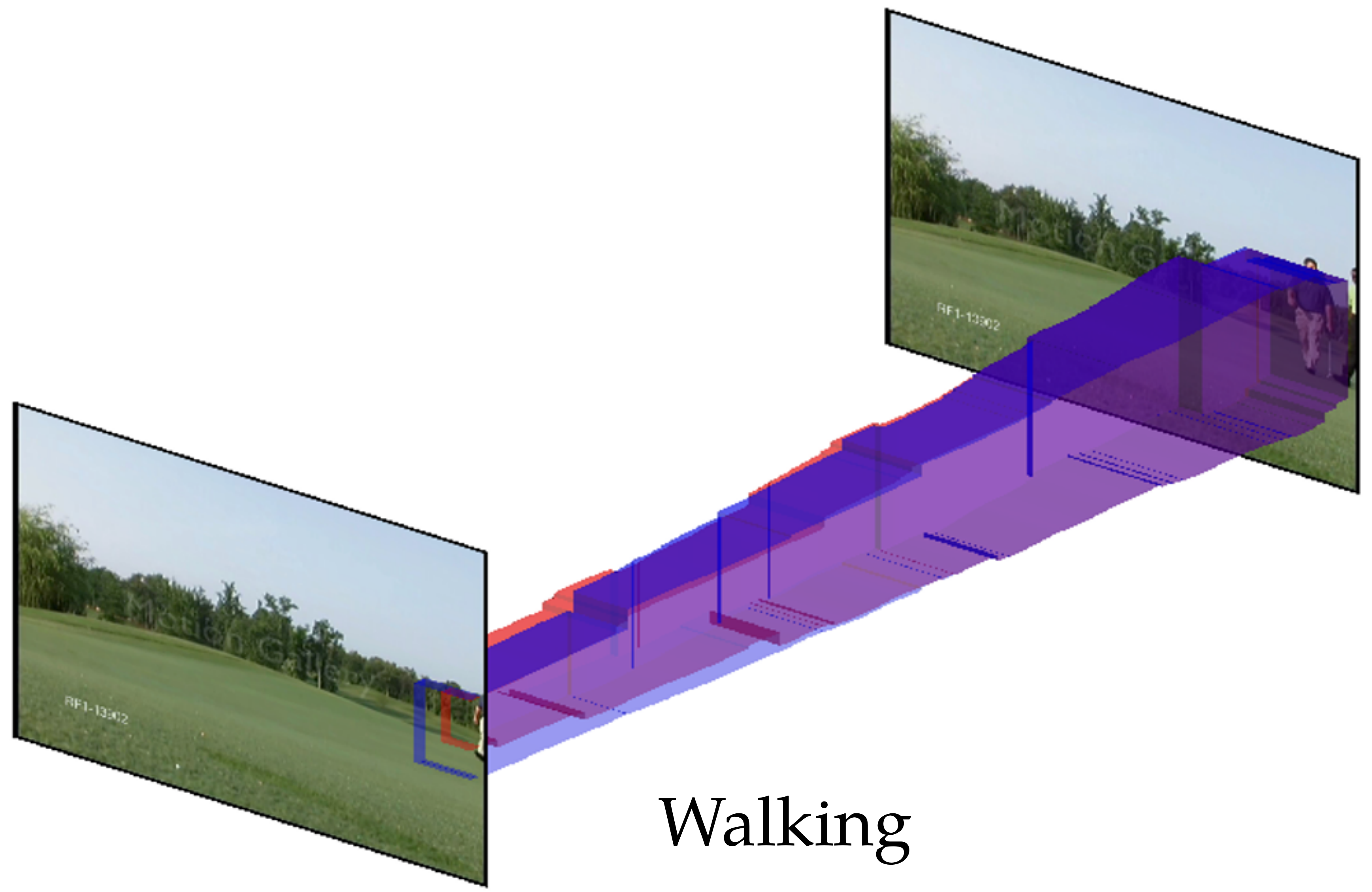}
\hspace{0.5cm}
\includegraphics[width=0.3\textwidth]{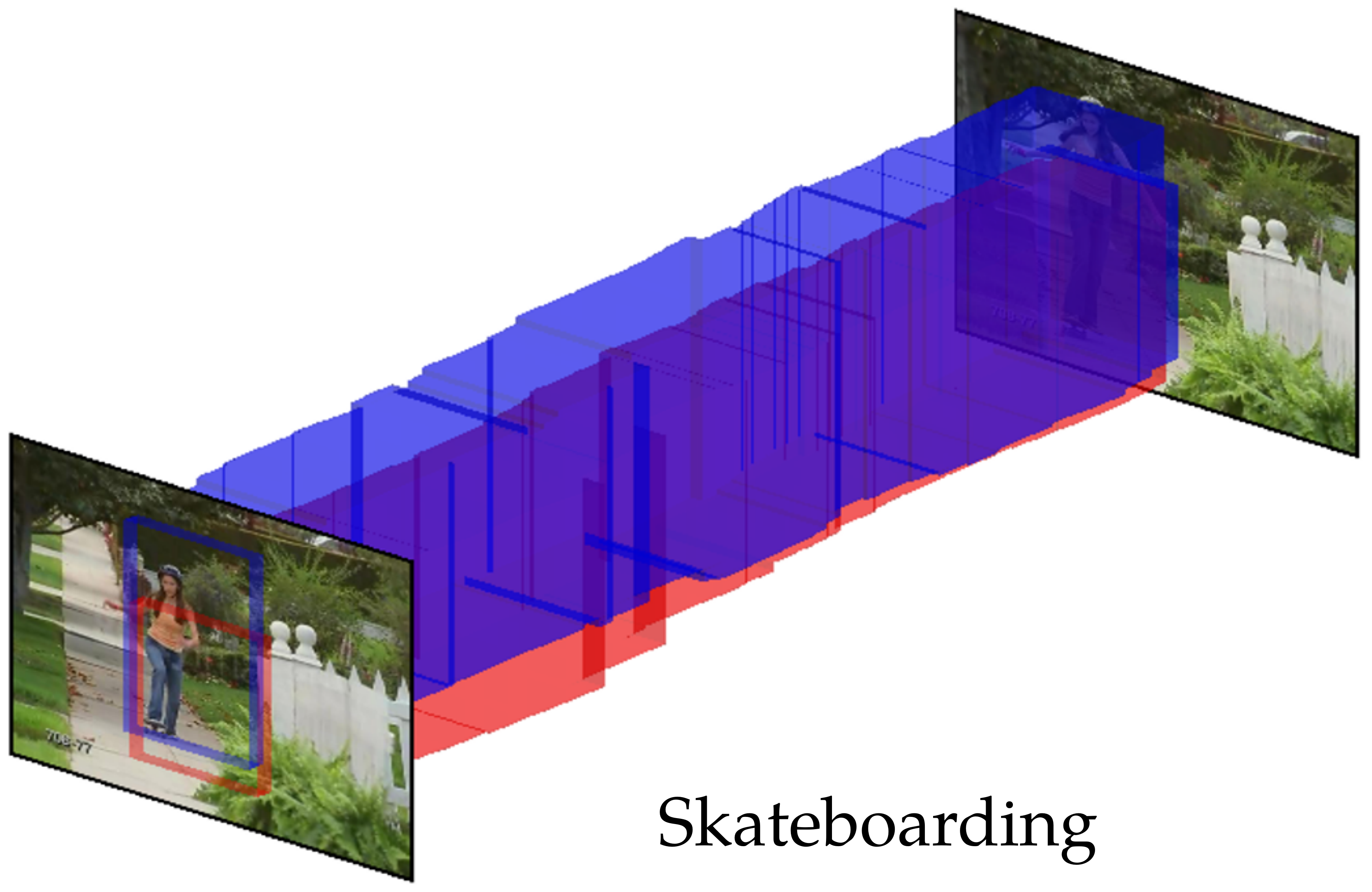}
\hspace{0.5cm}
\includegraphics[width=0.3\textwidth]{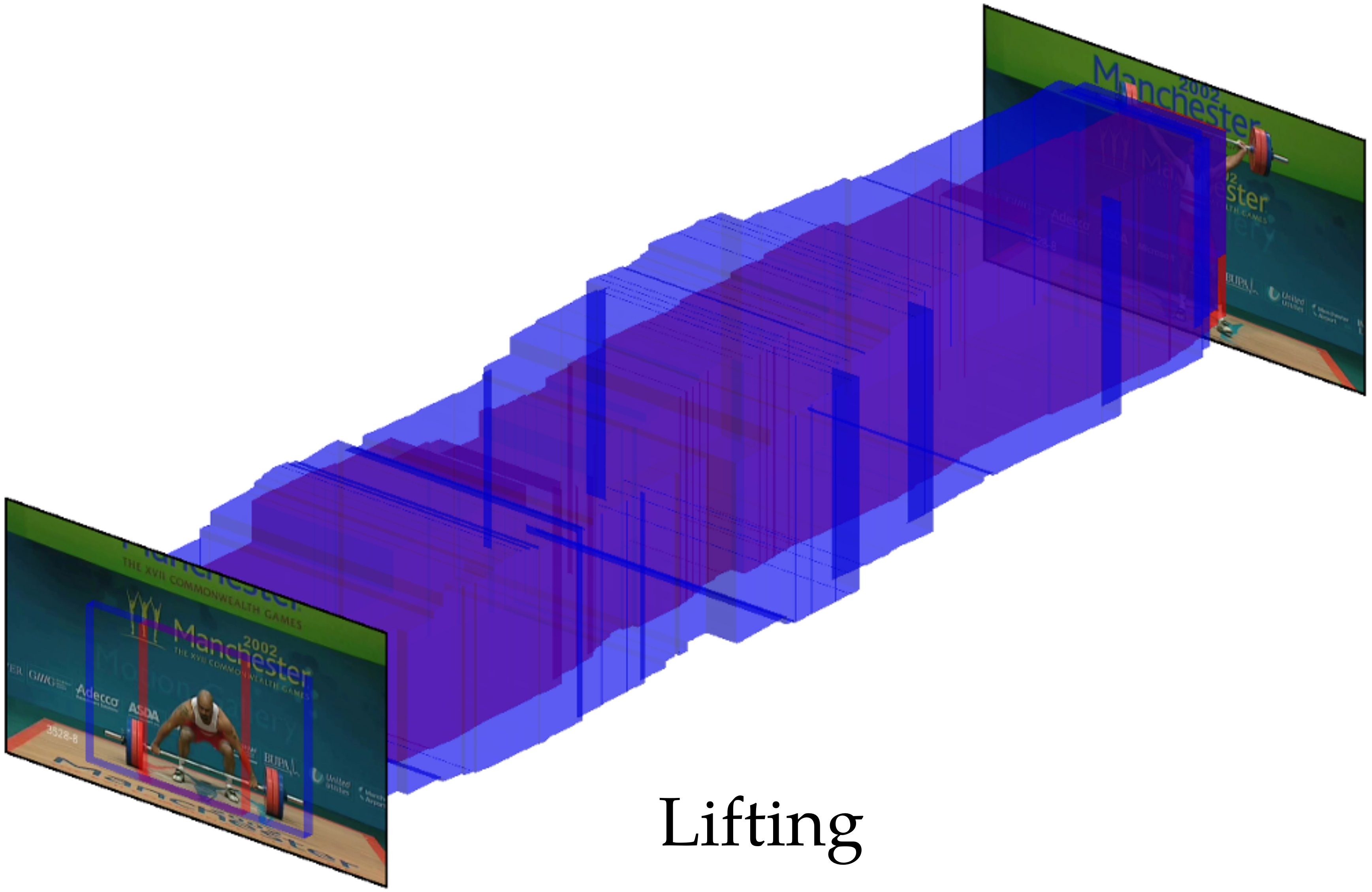}
\caption{}
\end{subfigure}
\begin{subfigure}{\textwidth}
\centering
\includegraphics[width=0.3\textwidth]{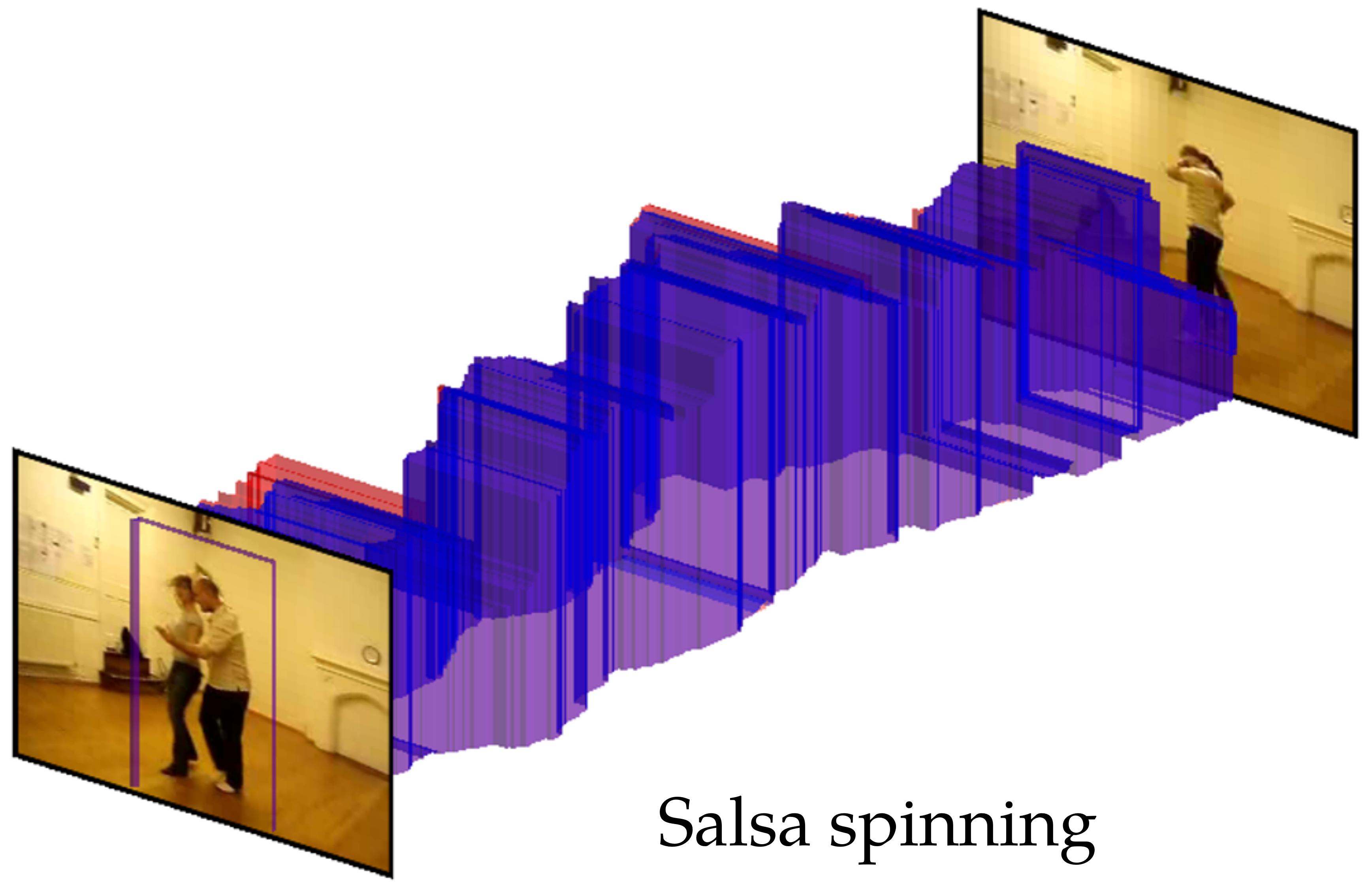}
\hspace{0.5cm}
\includegraphics[width=0.3\textwidth]{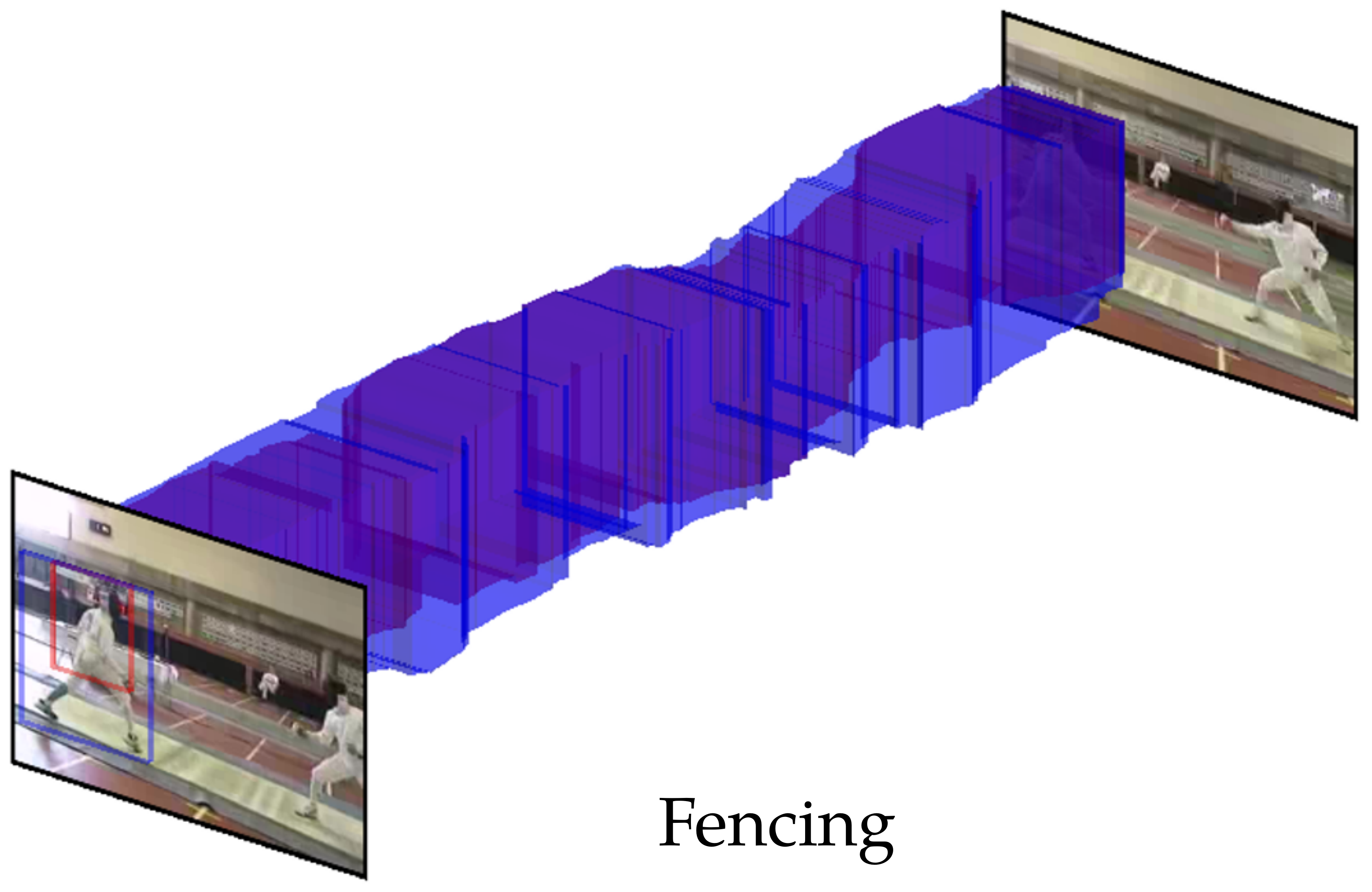}
\hspace{0.5cm}
\includegraphics[width=0.3\textwidth]{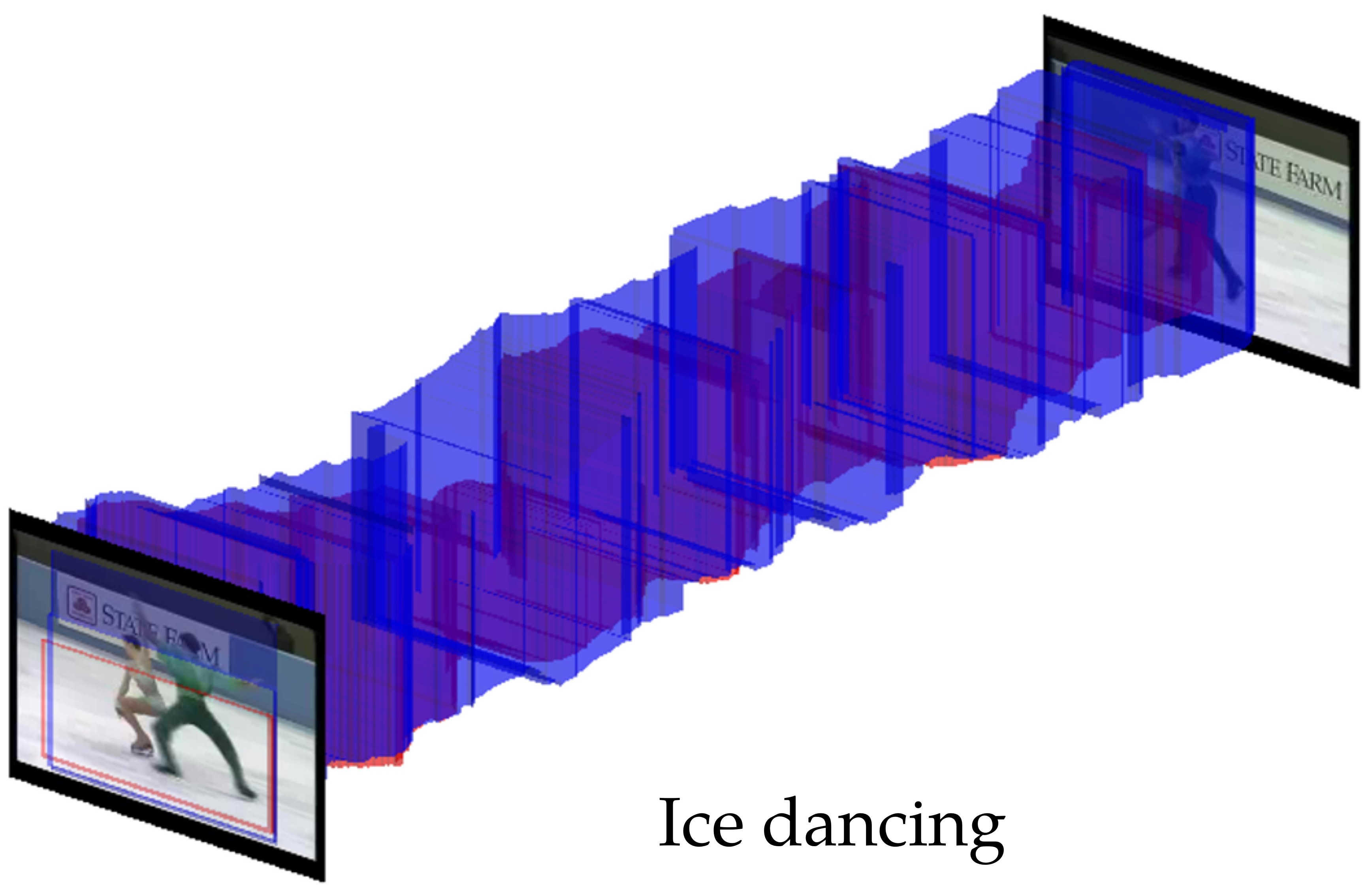}
\\
\includegraphics[width=0.3\textwidth]{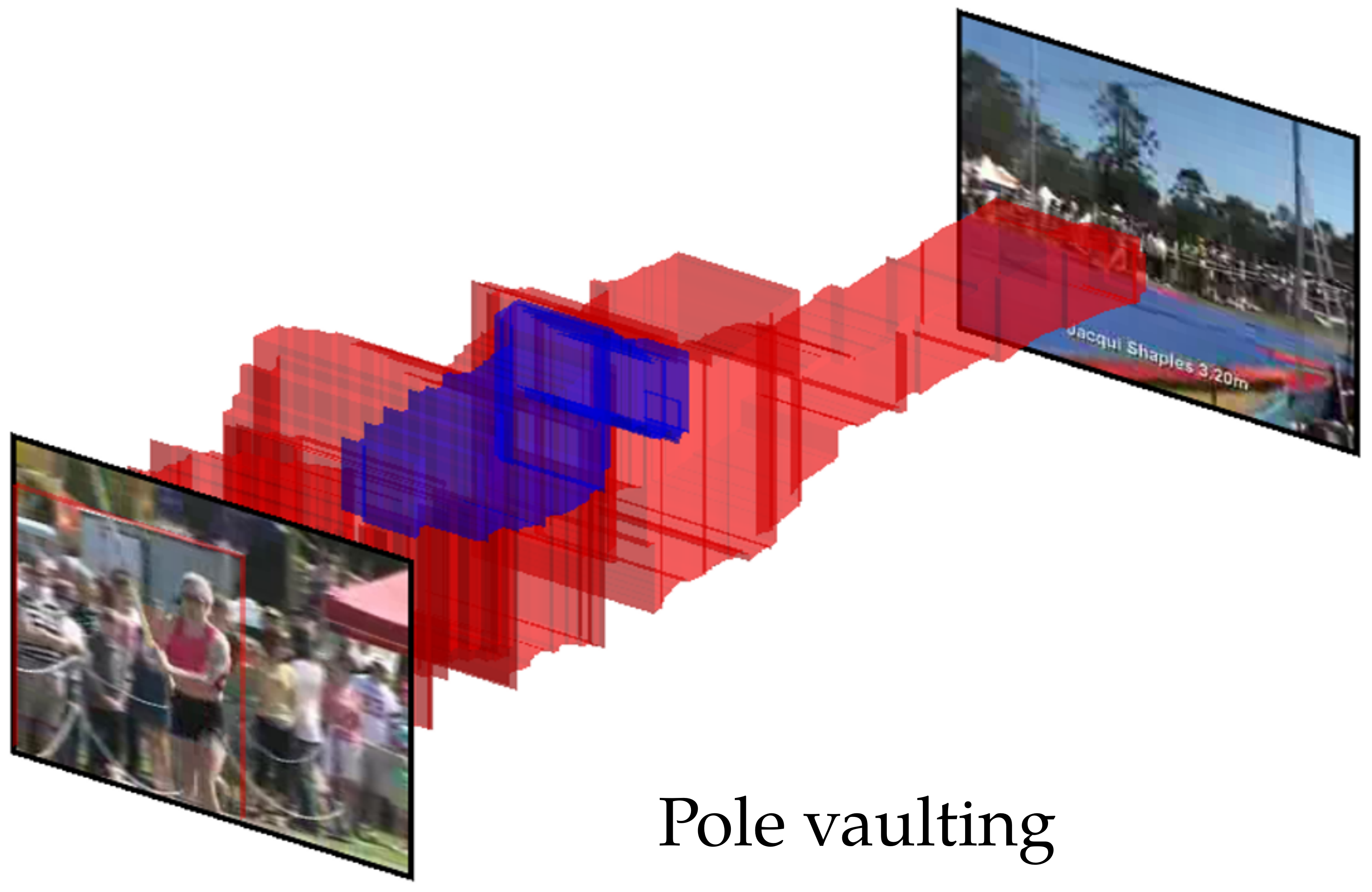}
\hspace{0.5cm}
\includegraphics[width=0.3\textwidth]{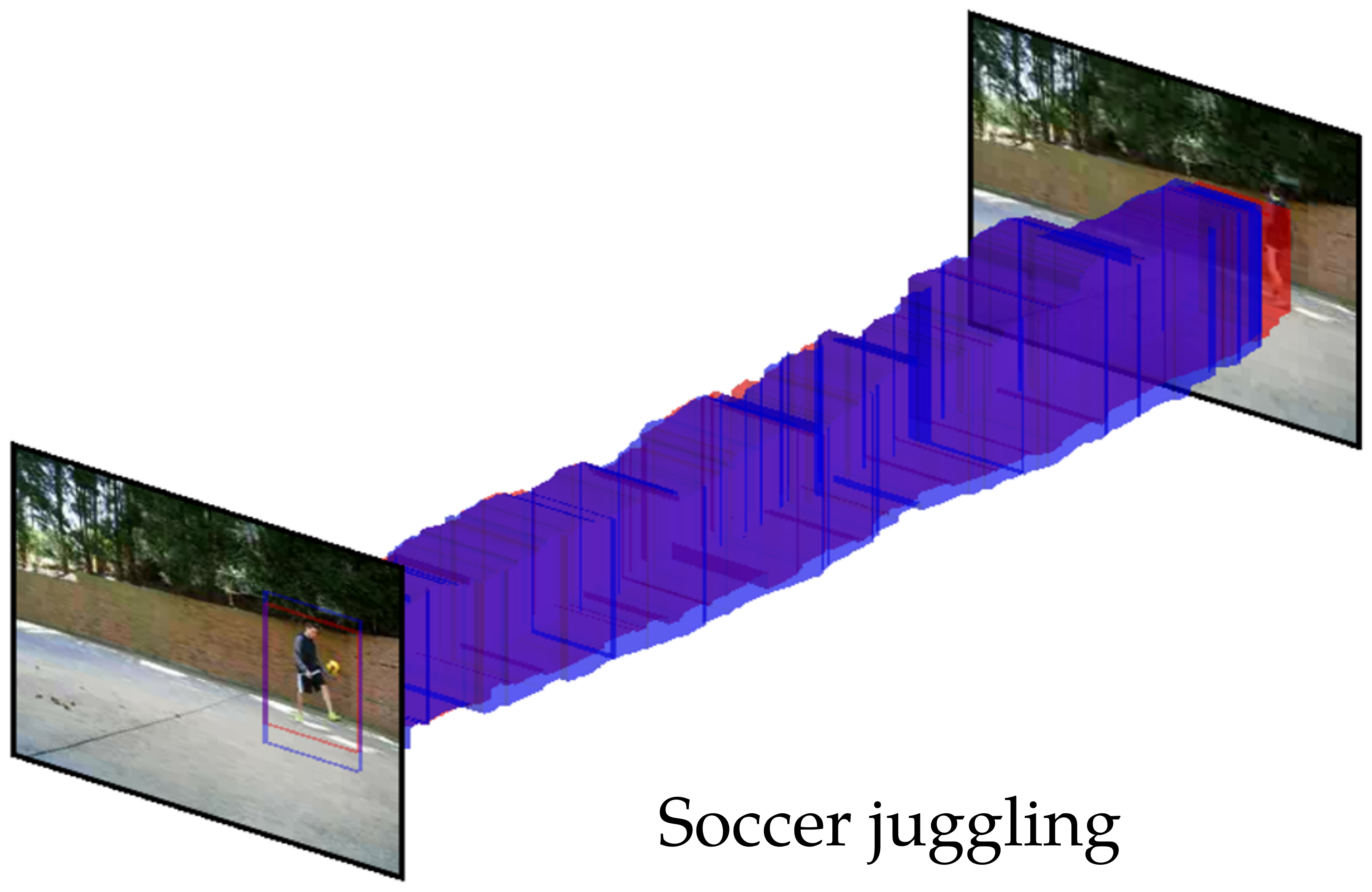}
\hspace{0.5cm}
\includegraphics[width=0.3\textwidth]{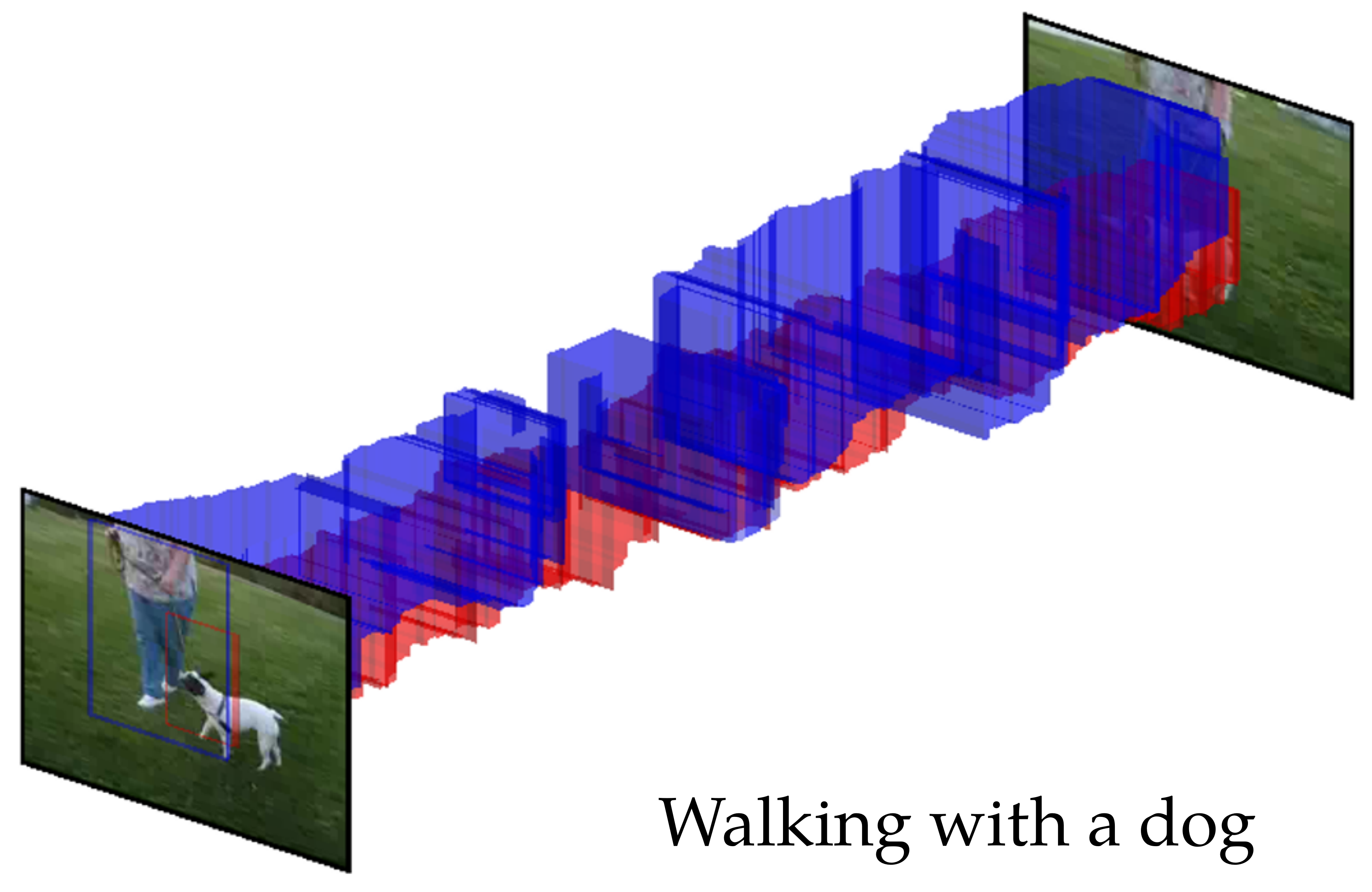}
\caption{}
\label{fig:exp1-qual-b}
\end{subfigure}
\caption{Qualitative results on (a) UCF Sports and (b) UCF-101 of selected proposals using point- (red) and box-supervision (blue).
For simple actions with static backgrounds, such as \emph{walking}, \emph{salsa spinning} and \emph{soccer juggling}, both approaches converge to similar locations. For actions with a more dynamic background and interacting objects, point-supervision might lead to a selection of different proposal locations. Examples include \emph{fencing} and \emph{walking with a dog}. We conclude that point- and box-supervision typically leads to similar results, although point-supervision tends to focus more on the most discriminative element of actions.}
\label{fig:exp1-qual}
\end{figure*}
\textbf{Error analysis.}
To gain insight into why point supervision is effective for action localization, we perform an error diagnosis and corresponding qualitative analysis. We perform the diagnosis on the approaches using box-supervision and point-supervision. Akin to error diagnosis for object detection~\citep{hoiem2012diagnosing}, we quantify the types of errors made by each localization approach. We take the top $R$ detections for each action, where $R$ is equal to the number of ground truth instances in the test set. We categorize each detection into five classes relevant for action localization: (1) correct detection, (2) localization error, (3) confusion with other action, (4) background from video containing the action, and (5) background from video not containing the action. The categorization definition is provided in Appendix~\ref{sec:appendix}.

The error diagnosis, averaged over all actions, is shown in Figure~\ref{fig:exp1-error-avg} for UCF Sports and UCF-101.
We observe that overall, the types of errors made by both approaches are similar. The predominant error type is localization error, which means that proposals from positive videos with a low match to the ground truth are the main errors. Proposals from background proposals of both positive and negative videos are hardly ranked high. Overall, using boxes and points result in similar errors, which matches with their similarity in localization performance. A common limitation is the quality of the spatio-temporal proposals themselves; only few proposals have a high overlap with the ground truth, making the localization a needle in the haystack problem regardless of the model.
On UCF-101, a large part of the errors also comes from confusion with the background from other videos. This is because the UCF-101 dataset can have more than one instance of the same action in each video. If such additional instances are missed, non-distinct regions of negative videos are automatically ranked higher.
%
\\\\
\textbf{Qualitative analysis.}
In Figure~\ref{fig:exp1-qual}, we provide qualitative results on UCF Sports and UCF-101.
The results show where point- and box-supervision yield similar and dissimilar action tubes. For simple actions like \emph{walking} and \emph{soccer juggling}, both approaches yield (near-)identical results. For actions such as \emph{skateboarding} and \emph{walking with a dog}, we observe that point-supervision tends to focus on the invariant object (here: skateboard, horse, and dog), since these are distinctive elements for the action. This is because the spatial extent of actions is no longer known with points, which means that the extent is learned from examples, rather than from manual annotations. We also note that limitations in the model can result in different results, as shown in the leftmost example of Figure~\ref{fig:exp1-qual-b}.
\\\\
\textbf{Conclusions.}
From the localization results, error diagnosis, and qualitative analysis, we make the following conclusions:
(i) point-supervision yield results comparable to full box-supervision for action localization,
(ii) averaged over all actions, the approaches using box and point annotations have approximately similar error type distributions,and
(iii) models learned with point-supervision learn the spatial extent of actions discriminatively from examples.

\begin{table*}[t]
\centering
\begin{tabular}{lcccccccc}
\toprule
 & \textbf{Box supervision} & \multicolumn{7}{c}{\textbf{Point supervision}}\\
 &  & \multicolumn{7}{c}{Annotation stride}\\
 & & 1 & 2 & 5 & 10 & 20 & 50 & 100\\
\midrule
mAP@0.2 & 0.399 & 0.393 & 0.404 & 0.389 & 0.384 & 0.395 & 0.379 & 0.371\\
mAP@0.5 & 0.074 & 0.063 & 0.060 & 0.068 & 0.064 & 0.061 & 0.064 & 0.053\\
\rowcolor{Gray}
Annotation speed-up & \textbf{1.0} & \textbf{9.8} & \textbf{19.3} & \textbf{46.0} & \textbf{85.0} & \textbf{147.6} & \textbf{264.6} & \textbf{359.6}\\
\bottomrule
\end{tabular}
\caption{Action localization performance on UCF-101 as a function of the annotation stride for point-supervision, compared to box-supervision. The annotation-speedup is relative to a box annotation in each frame. Fewer point annotations result in large annotation-time speed-ups, while the performance is hardly affected.
}
\label{tab:exp2}
\end{table*}

\begin{figure}[t]
\includegraphics[width=\linewidth]{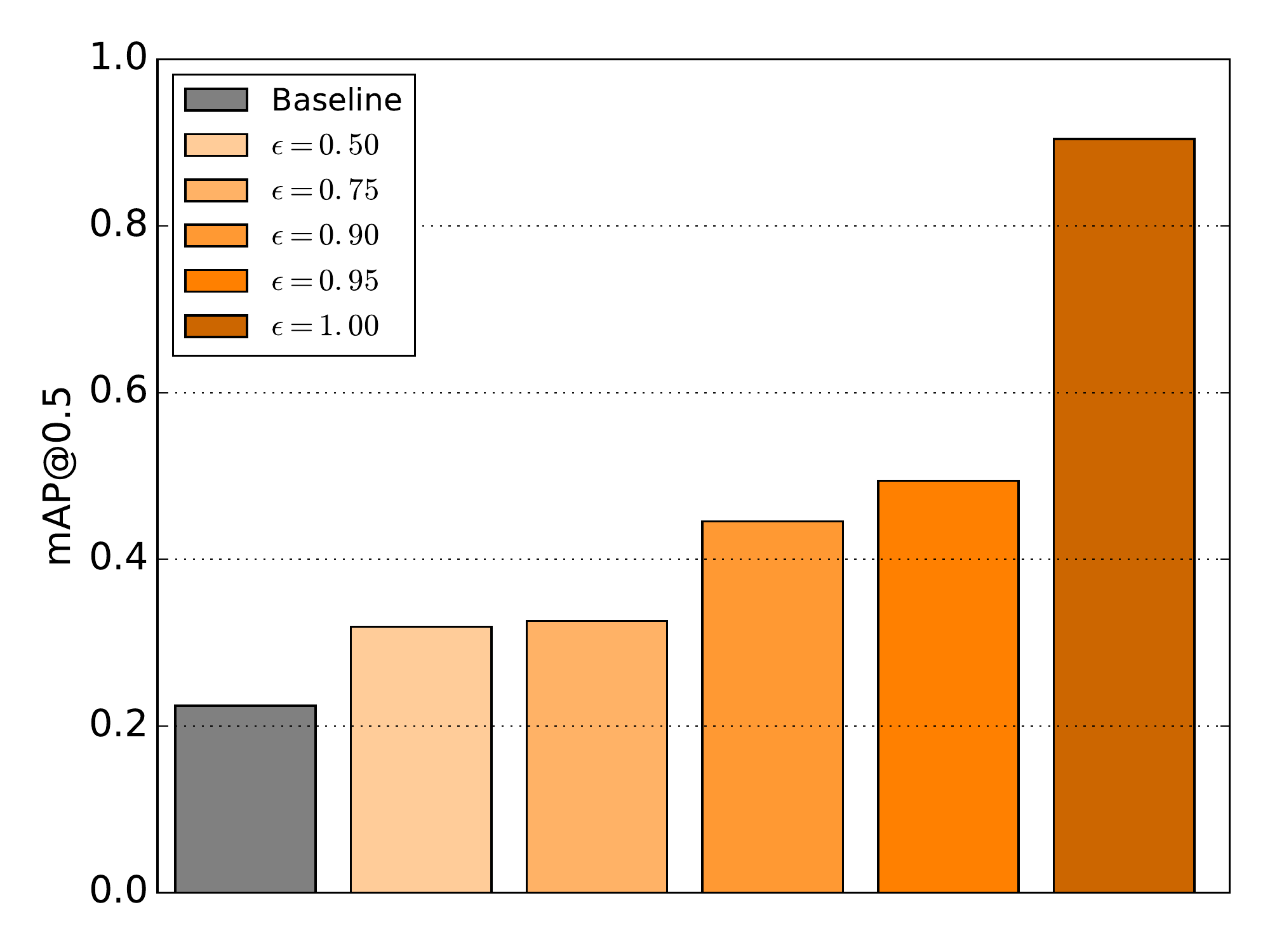}
\caption{Influence of spatio-temporal proposal quality on UCF-Sports. The baseline corresponds to the result from the first experiment. For the others, the ground truth location is added as one of the proposals during testing. Where, $\epsilon$ states the fraction of low quality (overlap $\leq$ 0.5) proposals that are removed. Action localization performance increases when large amounts of low quality proposals are removed. We conclude that better quality action proposals will have a positive impact on pointly-supervised action localization.}
\label{fig:exp5}
\end{figure}

%
%
\subsection{Influence of spatio-temporal proposal quality}
In the second experiment, we evaluate the influence of the spatio-temporal proposals upon which our approach is built. Spatio-temporal proposals optimize recall, \ie for a video, \emph{at least} one proposal should have a high overlap to the ground truth action localization. An inconvenient side-effect from this requirement is that each video outputs many proposals that have a low overlap, making the selection of the best proposal during testing a needle in the haystack problem. This problem was observed in the error diagnosis of the first experiment.

Here, we investigate the influence of the high ratio of proposals with a low overlap during training and testing. During both training and testing, we add the oracle ground truth tube to the proposals. We furthermore add a parameter $\epsilon$, which controls the fraction of proposals with an overlap below 0.5. We train several models with varying values for $\epsilon$. We evaluate this oracle experiment on UCF Sports.
\\\\
\textbf{Results.}
The localization performance for several values of $\epsilon$ is shown in Figure~\ref{fig:exp5}. The baseline is the result achieved in the first experiment. From the Figure, it is evident that removing low quality proposals positively affects the localization performance. However, a large portion of low quality proposals need to be removed to achieve better results. This is because of the large amount of low quality proposals. On UCF Sports, only 7\% of the proposals have an overlap of at least 0.5 (!). This means that when 50\% of the low quality proposals are removed, the ratio of low to high quality proposals is still 6 to 1. When removing 50\% of the low quality proposals, the result increases from 0.23 to 0.32. This further increases to 0.49 when removing 95\% of the low quality proposals.
%
When only using the ground truth tube and high overlapping proposals (\ie $\epsilon=1.0$), we achieve a performance of 0.90, indicating the large gap between current performance and the upper bound given the current set of features.
%
We conclude from this experiment that for pointly-supervised action localization with spatio-temporal proposals, a limiting factor is the quality of the proposals themselves. With better action proposals, point-supervision can achieve even better results.

%
%
\subsection{Sparse point annotations}
In the third experiment, we evaluate: (i) how much faster is point-supervision compared to box-supervision and (ii) how many point annotations are sufficient for effective localization. Intuitively, point-supervision is not required for every frame, since the amount of change between consecutive frames is small.
We evaluate the influence of the annotation stride and we also estimate how much faster the annotation process becomes compared to dense box-supervision.
We perform this experiment on UCF-101, since the videos in this dataset are the longest for action localization, allowing for a wide range of annotation strides.
\\\\
\textbf{Annotation times.}
To obtain an estimate of the annotation times for box- and point-supervision, we have re-annotated several hundreds of videos while keeping track of the annotation times. We found that annotating a video with an action label takes roughly 5 seconds. Furthermore, annotating a box in a frame takes roughly 15 seconds. This estimate is in between the estimate for image annotation of \cite{su2012crowdsourcing} (roughly 30 seconds) and the estimate of \cite{russakovsky2015best} (10 to 12 seconds). Annotating a point takes roughly 1.5 seconds, making points ten times faster than boxes to annotate. This estimate is in line with point annotations in images (0.9-2.4 seconds~\citep{bearman2016s}).
\\\\
\textbf{Results.}
In Table~\ref{tab:exp2}, we provide the localization performance for two overlap thresholds and the annotation speed-up for point supervision at seven annotation strides. The Table shows that when annotating fewer frames, performance is retained. Only when annotating fewer than 5\% of the frames (\ie for a stride larger than 20), the performance drops marginally. This result shows that our approach is robust to sparse annotation, a point at every frame is not required. The bottom row of the Table shows the corresponding speed-up in annotation time compared to box-supervision. An almost 50-fold speed-up can be achieved while maintaining comparable localization performance. A 300-500 fold speed-up can be attained with a marginal drop in localization performance. We conclude that point-supervision is robust to sparse annotations, opening up the possibility for further reductions in annotation cost for action localization.

%
%
\subsection{Noisy point annotations}
\begin{figure}[t]
\includegraphics[width=\linewidth]{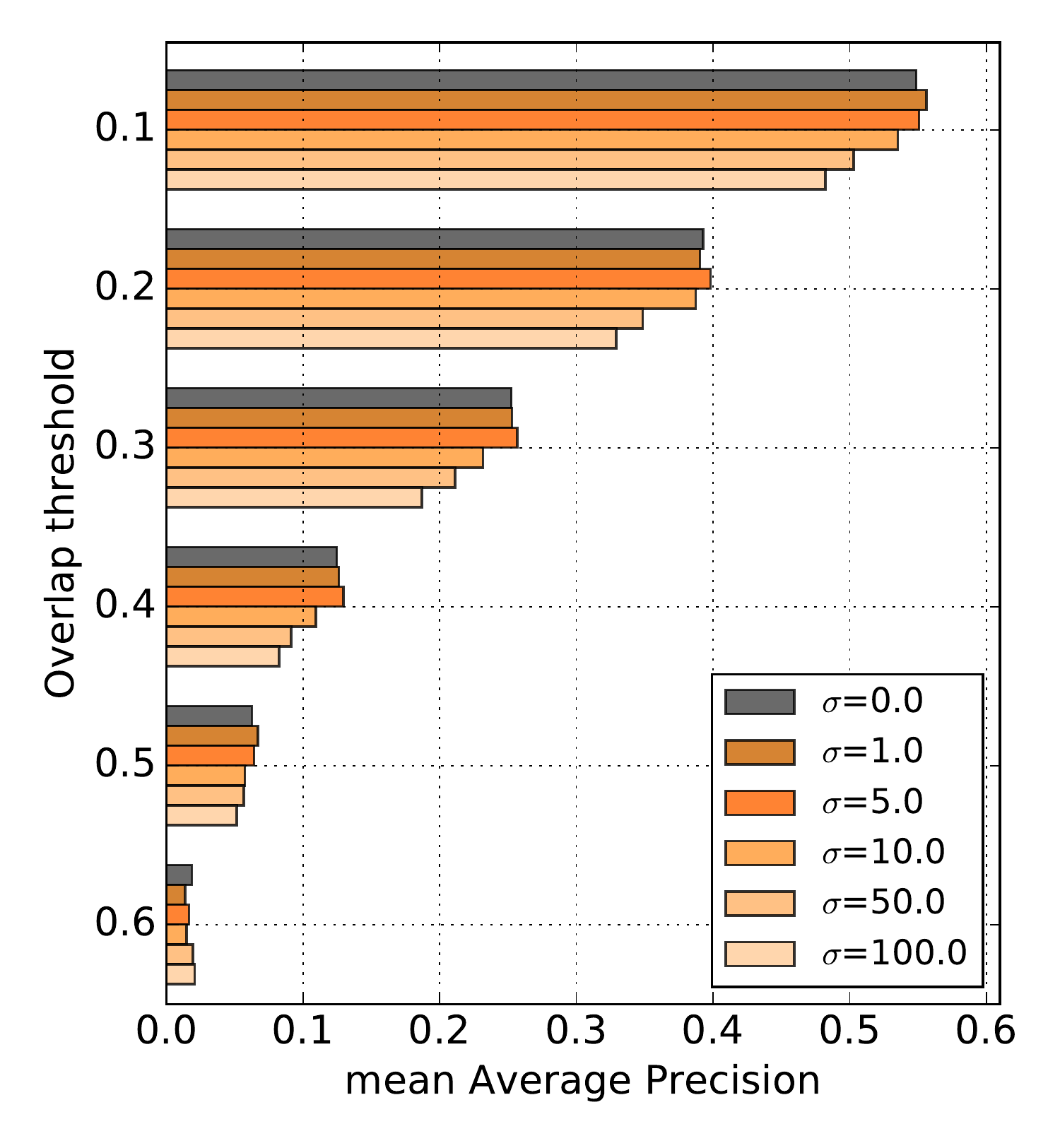}
\caption{Localization performance on UCF-101 for various levels of noise in the point annotations. Up to a noise deviation of 10 pixels can be handled robustly. For large deviations (50 pixels an up), performance drops for lower overlap thresholds. Point-supervision can accommodate human error in the point annotations up to 10 pixels.}
\label{fig:exp3-noise}
\end{figure}

Human annotators, while center biased~\citep{tseng2009quantifying}, do not always precisely pinpoint center locations while annotating~\citep{bearman2016s}.
In the fourth experiment, we evaluate how robust the action localization performance is with respect to noise in the point-supervision.
We start from the original point annotations and add zero-mean Gaussian noise with varying levels of isotropic variance.
This experiment is performed on the UCF-101 dataset.
\\\\
\textbf{Results.}
The localization performance for six levels of annotation noise is shown in Figure~\ref{fig:exp3-noise}. The performance for $\sigma=0$ corresponds to the performance of point-supervision in the first experiment. We observe that across all overlap thresholds, the localization performance is unaffected for noise variations up to a $\sigma$ of 5. For $\sigma=10$, the results are only affected for thresholds of 0.3 and 0.4, highlighting the robustness of our approach to annotation noise. For large noise variations ($\sigma=50$ or 100), we observe a modest drop in performance for the overlaps thresholds 0.1 to 0.4. We conclude that points do not need to be annotated precisely at the center of actions. Annotating points in the vicinity of the action is sufficient for action localization.

\begin{figure*}[t]
\centering
\begin{subfigure}{\textwidth}
\centering
\includegraphics[width=0.49\linewidth]{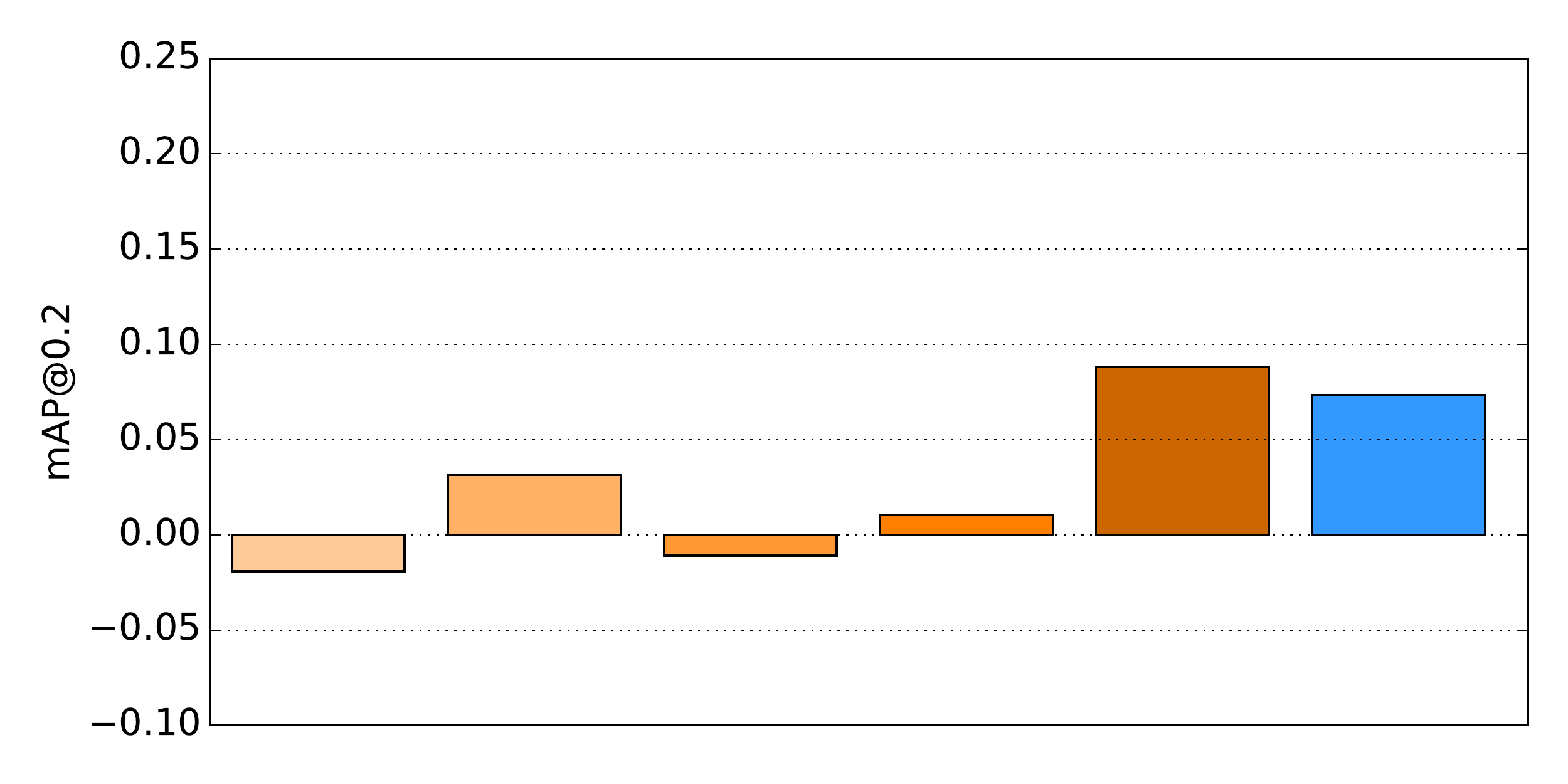}
\includegraphics[width=0.49\linewidth]{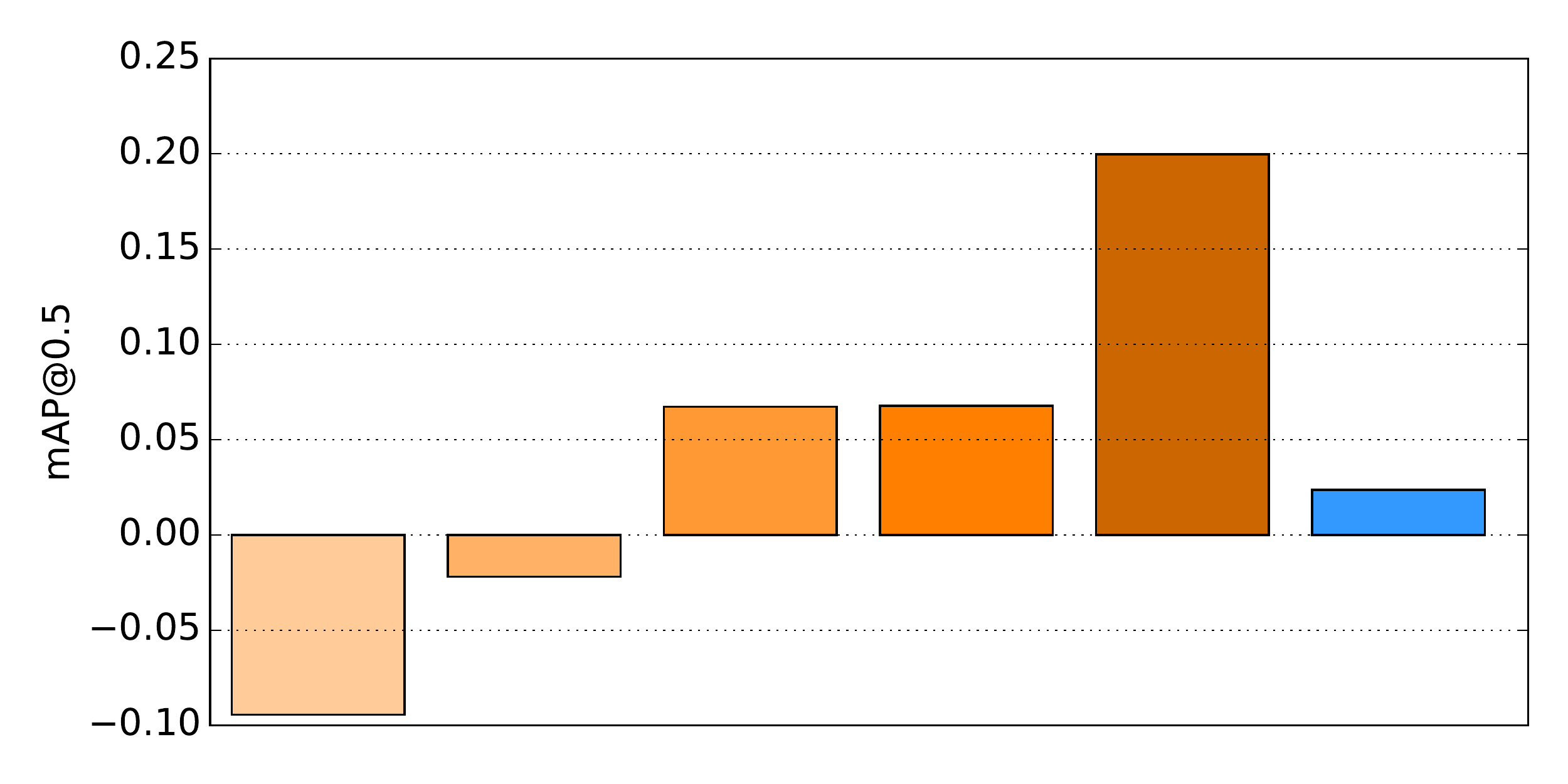}
\caption{UCF Sports.}
\label{fig:exp4-ucfsports}
\end{subfigure}
\begin{subfigure}{\textwidth}
\centering
\includegraphics[width=0.49\linewidth]{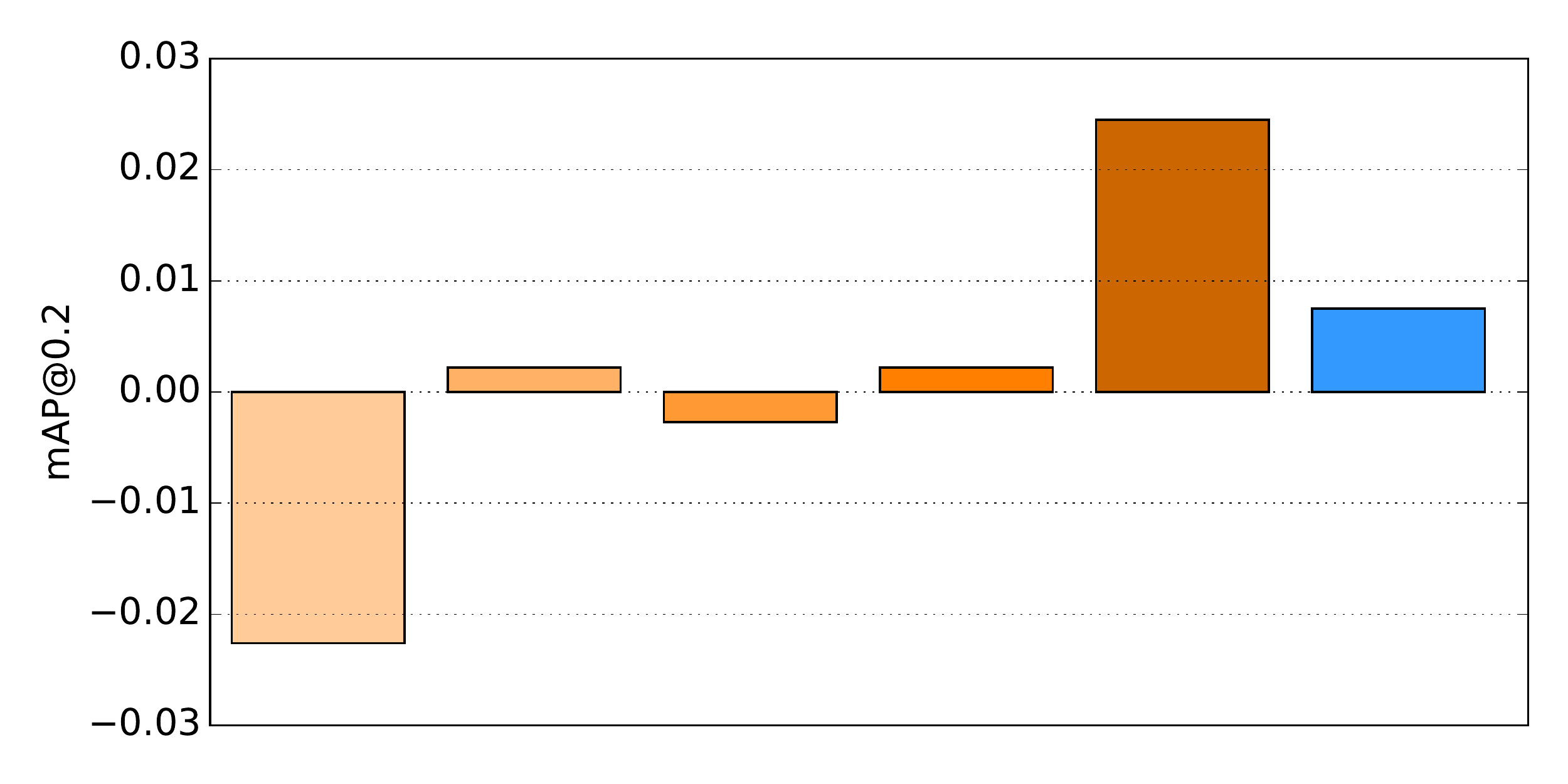}
\includegraphics[width=0.49\linewidth]{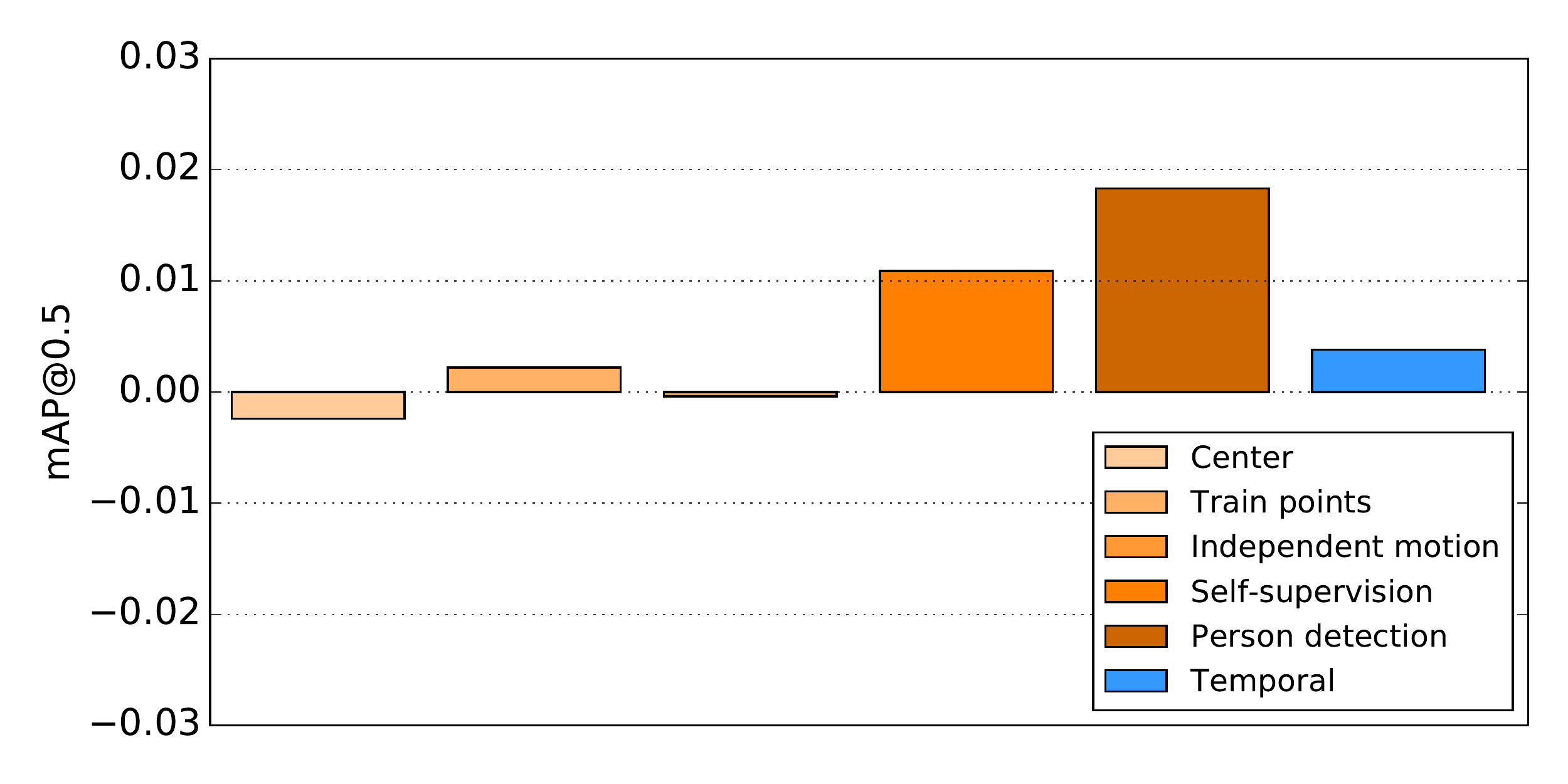}
\caption{UCF-101.}
\label{fig:exp4-ucf101}
\end{subfigure}
\caption{The effect of pseudo-points during inference for action localization on UCF Sports and UCF-101 at an overlap of 0.2 (left column) and 0.5 (right column). All results are provided relative to the performance without pseudo-points. Data-dependent pseudo-points such as person detection, self-supervision, and independent motion have a positive effect on the localization performance. Data-independent pseudo-points such as center bias and training points are not effective for action localization. Incorporating the temporal extent of actions as a pseudo-point can further boost performance. We conclude that pseudo-points, when chosen correctly, aid action localization performance.}
\label{fig:exp4}
\end{figure*}

%
%
\subsection{Exploiting pseudo-points}
In the fifth experiment, we investigate the effect of each of the pseudo-points on the action localization performance during inference. We perform this experiment on both UCF Sports and UCF-101.
\\\\
\textbf{Pseudo-point weights.}
To utilize the pseudo-points effectively during inference and to know a priori which pseudo-point is most effective, we compute the weight per pseduo-point as outlined in Section~\ref{sec:pa-weights}. This has resulted in the following values:
\begin{enumerate}
\item Person detection: $\lambda_P =$ 0.76.
\item Independent motion: $\lambda_P =$ 0.57.
\item Center bias: $\lambda_P =$ 0.48.
\item Self-supervision: $\lambda_P =$ 0.32.
\item Training points: $\lambda_P =$ 0.25.
\end{enumerate}
The weights computed based on the match with point-supervision in training videos provide the degree to which each pseudo-point should contribute to the selection of spatio-temporal proposals in test videos and they also provide a measure to select the best pseudo-point.
\\\\
\textbf{Results.}
In Figure~\ref{fig:exp4}, we show the localization performance for overlap thresholds of 0.2 and 0.5. On UCF Sports, we observe for an overlap of 0.2, the performance improves for training points, self-supervision, and person detection. For center bias and independent motion, there is a minimal drop in performance. For an overlap of 0.5, the results diverge more clearly. Independent motion (+6.7\%), self-supervision (+6.8\%), and person detection (+20.0\%) benefit directly from inclusion. This does not hold for center bias and training points. While pseudo-points can have a positive impact on the performance, it is not effective for all types of pseudo-points. Discovering which pseudo-points are most effective is a necessity.
On UCF-101, we observe similar trends. Person detection and self-supervision yield increased localization performance, while the data independent center bias and train points have a negative effect. Video-specific visual cues, such as persons and motion, are effective; generic statistics less so.

We observe that the order of the weights correlates with the localization performance. This indicates the effectiveness of the proposed weighting function, as it provides insight into the quality of the pseudo-points without having to evaluate their performance at test time. Person detection is the most effective pseudo-point. The center bias and training points score lower, which is also visible in their localization performance. Only the self-supervision scores low, while it has a positive effect on the localization. We conclude that the proposed pseudo-point weighting is a reliable way to determine the effectiveness of pseudo-points with point-supervision.

On both datasets, we also investigate the effect of temporal pseudo-points on the localization performance. On UCF Sports, we observe an increase in performance for both overlap thresholds (+7.3\% at 0.2, +2.4\% at 0.5). On UCF-101, we also observe a positive effect, albeit with smaller improvements (+0.8\% at 0.2, +0.6\% at 0.5). We conclude that regularizing spatio-temporal proposals using the temporal extent of actions, which is provided by point-supervision, aids action localization in videos.

Based on the weights of the pseudo-annotations, we recommend to use person detection during inference. We will use this setup for the state-of-the-art comparison.

\begin{figure*}[t]
\centering
\begin{subfigure}{\textwidth}
\centering
\includegraphics[width=0.32\linewidth]{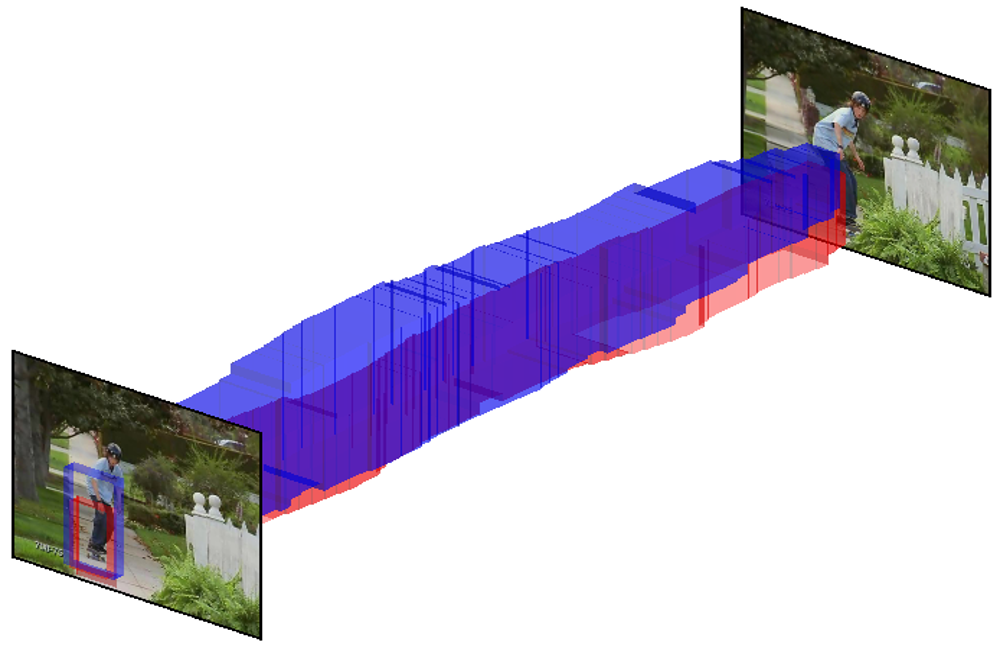}
\includegraphics[width=0.32\linewidth]{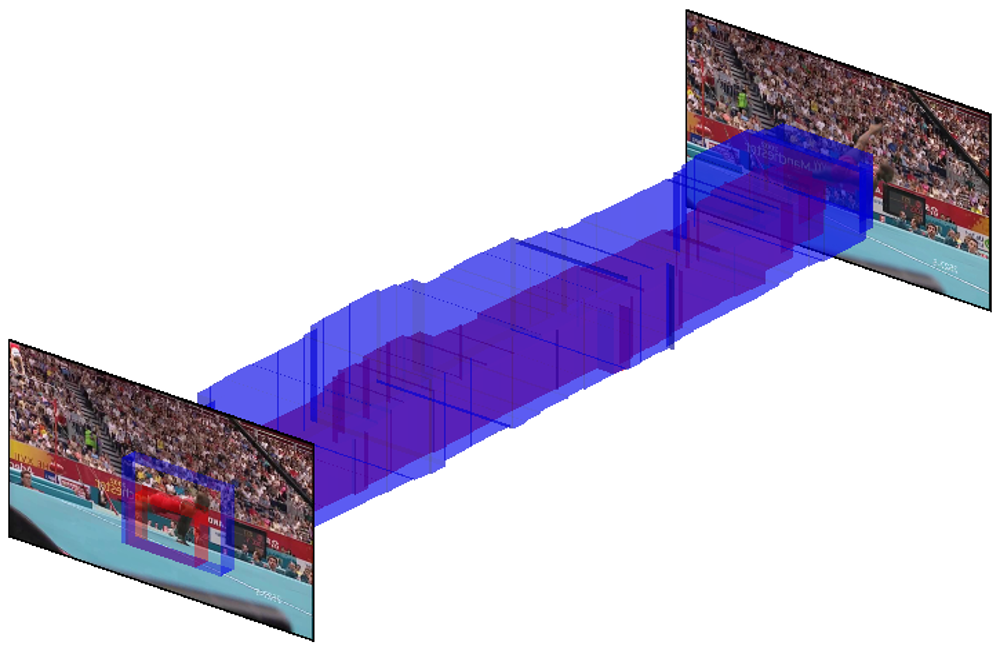}
\includegraphics[width=0.32\linewidth]{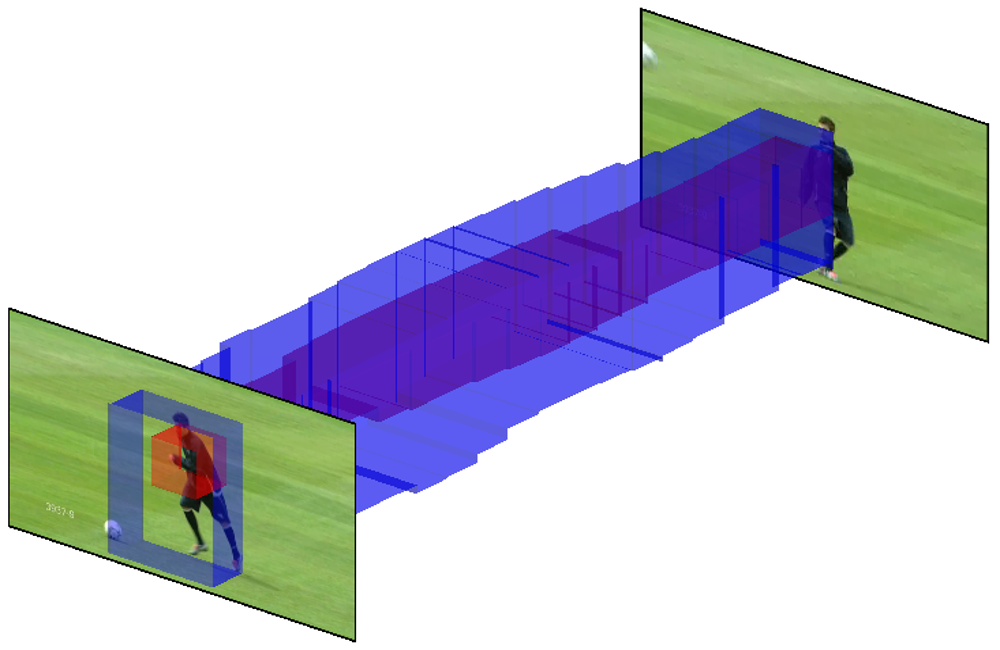}
\caption{Left to right: self-supervision, independent motion, and person detection.}
\label{fig:exp4-qual-a}
\end{subfigure}
\begin{subfigure}{\textwidth}
\centering
\includegraphics[width=0.32\linewidth]{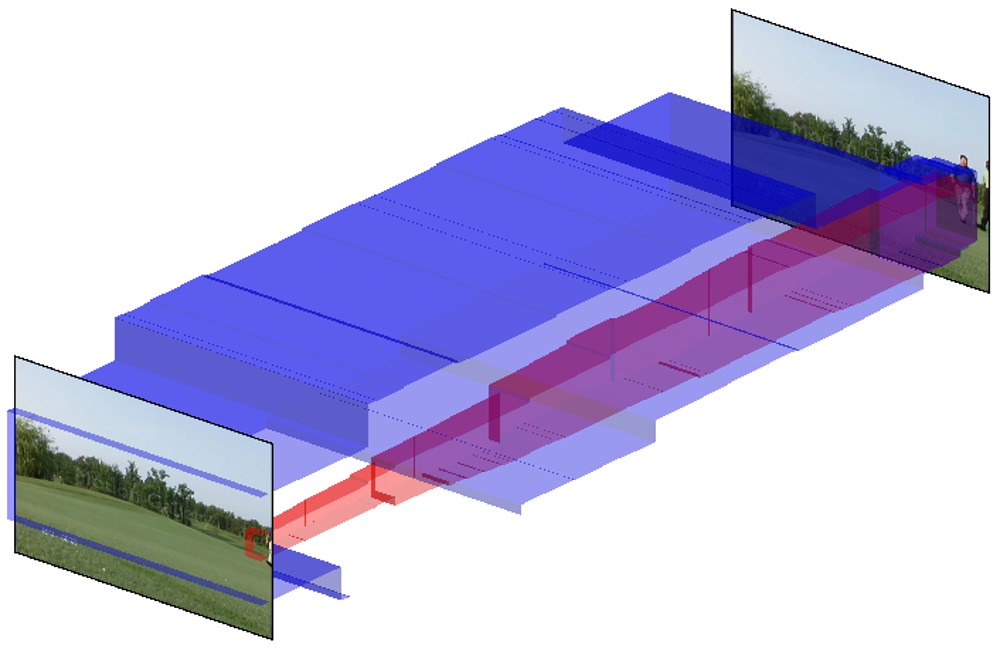}
\includegraphics[width=0.32\linewidth]{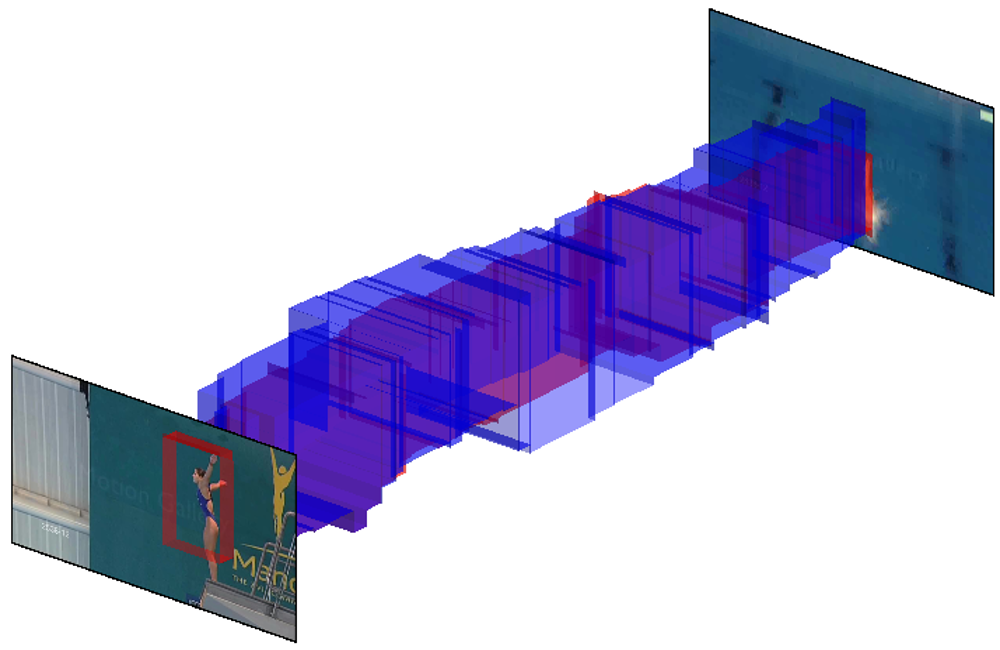}
\includegraphics[width=0.32\linewidth]{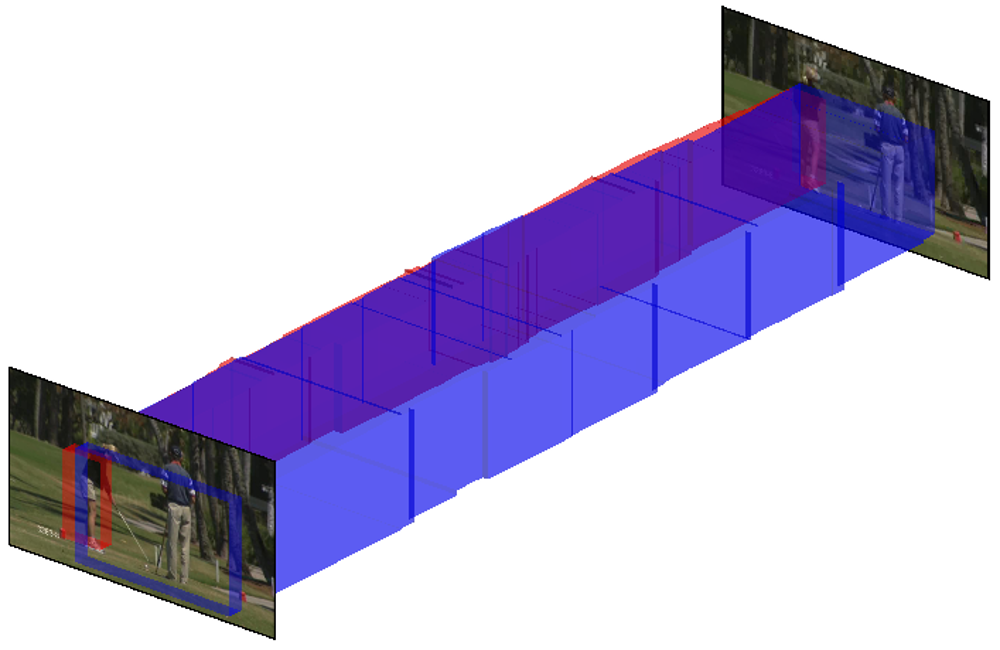}
\caption{Left to right: center bias, training points, and person detection.}
\label{fig:exp4-qual-b}
\end{subfigure}
\caption{Qualitative analysis of the effect of pseudo-points for action localization on UCF Sports. Red indicates the localization without pseudo-points, blue with pseudo-points. (a) The top row show three examples where pseudo-points improve the localization. For the self-supervision and independent motion examples, motion information helped to widen the scope of the action. For person detection, information from the whole person enhances the scope of the action. (b) The second row shows failure cases. For data independent pseudo-points such as center bias and training point statistics, the action can deviate from its true location. Person detection can furthermore be problematic when many people are present in the scene. The qualitative analysis shows that pseudo-points can alleviate problems in pointly-supervised action localization. To aid the performance, data-dependent pseudo-points are informative, while data-independent pseudo-points appear less effective.}
\label{fig:exp4-qual}
\end{figure*}

\textbf{Qualitative results.}
To gain insight into which types of videos benefit from pseudo-points, we provide a qualitative analysis on UCF Sports. In Figure~\ref{fig:exp4-qual-a}, we show three test videos where the localization improved due to the effect of pseudo-points. We show the effect for the independent motion, self-supervision, and person detection respectively. In all three cases, the inclusion of pseudo-points resulted in a better fit on the action by enlarging its scope. For self-supervision and independent motion, the wider motion evidence resulted in a better fitting localization. For person detection, the evidence of the whole person had a positive effect. These examples show the potential of pseudo-points to guide the selection of spatio-temporal proposals during inference.

In Figure~\ref{fig:exp4-qual-b}, we show three test videos where the inclusion of pseudo-points resulted in a worse localization. We show this effect for the less successful center bias and training points, as well as for the most successful pseudo-point person detection. For center bias, this resulted in a shift from precise fit on the action (red) to a large generic location (blue). This is because the center bias is data independent and might undo correct localizations. This also holds for the training point statistics, which are identical for each test video. Lastly, the person detection can yield diverging localizations when many people are present in the scene. We conclude that motion-based and person-based pseudo-points can aid action localization, while data independent pseudo-points are less suited.

\begin{table*}[t]
\centering
\scalebox{0.925}{
\begin{tabular}{lllccc}
\toprule
 & \textbf{Proposals} & \textbf{Supervision} & \textbf{UCF Sports} & \textbf{UCF-101} & \textbf{Hollywood2Tubes}\\
 & & & (AUC) & (mAP) & (mAP)\\
\midrule
\cite{lan2011discriminative} & $\times$ & box & 0.380 & - & -\\
\cite{TianPartCVPR2013} & $\times$ & box & 0.420 & - & -\\
\cite{wang2014video} & $\times$ & box & 0.470 & - & -\\
\cite{jain2014action} & $\checkmark$ & box & 0.520 & - & -\\
\cite{gemert2015apt} & $\checkmark$ & box & 0.546 & 0.345 & -\\
\cite{soomroICCV2015actionLocContextWalk} & $\checkmark$ & box & 0.550 & - & -\\
\cite{gkioxari2015finding} & $\times$ & box & 0.559 & - & -\\
\cite{weinzaepfelICCV2015learningToTrack} & $\times$ & box & 0.559 & 0.468 & -\\
\cite{jain2017tubelets} & $\checkmark$ & box & 0.570 & 0.475 & -\\
\cite{hou2017tube} & $\times$ & box & 0.580 & 0.471 & -\\
\cite{saha2017amtnet} & $\times$ & box & - & 0.631 & -\\
\cite{yang2017spatio} & $\times$ & box & - & 0.735 & -\\
\cite{kalogeiton2017action} & $\times$ & box & - & \textbf{0.772} & -\\
\midrule
\cite{jain2015objects2action} & $\checkmark$ & zero-shot & 0.232 & - & -\\
\cite{mettes2017spatial} & $\times$ & zero-shot & 0.393 & - & -\\
\cite{soomro2017unsupervised} & $\times$ & unsupervised & 0.450 & - & -\\
\cite{sharma2015action} from \citep{li2018videolstm} & $\times$ & video-label & - & 0.055 & -\\
\cite{cinbis2017weakly} from \citep{mettes2016spot} & $\checkmark$ & video-label & 0.278 & 0.136 & 0.009\\
\cite{chen2015action} & $\checkmark$ & video-label & 0.530 & - & -\\
\cite{li2018videolstm} & $\times$ & video-label & - & 0.369 & -\\
\cite{mettes2017localizing} & $\checkmark$ & video-label & 0.556 & 0.374 & 0.172\\
\cite{mettes2016spot} & $\checkmark$ & point & 0.545 & 0.348 & 0.143\\
\rowcolor{Gray}
\textbf{This paper} & $\checkmark$ & point & \textbf{0.598} & 0.418 & \textbf{0.178}\\
\bottomrule
\end{tabular}
}
\caption{Comparative evaluation of pointly-supervised action localization to the state-of-the-art using box-supervision as well as weakly-supervised alternatives. All results are shown for an overlap threshold of 0.2. On all datasets our approach compares favorably to all weakly-supervised localization approaches, indicating the effectiveness of point-supervision. On UCF Sports, we perform comparable or better to approaches that require box-supervision. On UCF-101, we outperform the approach based on box-supervision with the same proposals and features~\citep{gemert2015apt}, but we are outperformed by approaches that score and link individual boxes~\citep{kalogeiton2017action,saha2017amtnet,yang2017spatio}. We expect that higher quality spatio-temporal proposals, can narrow this gap (see Figure~\ref{fig:exp5}). On Hollywood2Tubes, which only provides point annotations for training, we set a new state-of-the-art.}
\label{tab:sota}
\end{table*}

%
%
\subsection{Comparison to others}
In our final experiment, we compare pointly-supervised action localization with alternatives using either box-supervision, or weaker forms of supervision. We perform this experiment on UCF Sports, UCF-101, and Hollywood2Tubes. To compare with as many methods as possible, we evaluate with AUC on UCF Sports and with mAP on UCF-101 and Hollywood2Tubes. We evaluate action localization for the standard overlap threshold of 0.2.
\\\\
\textbf{Results.}
%
%
We present results on all three datasets In Table~\ref{tab:sota}. On UCF Sports, we observe that our approach outperforms the state-of-the-art in weakly-supervised action localization, as well as the point-supervision in our previous work~\citep{mettes2016spot} which lacks the pseudo-points during inference.

Naturally, our pointly-supervised approach also outperforms the state-of-the-art in zero-shot and unsupervised action localization, emphasizing the effectiveness of points as supervision. Lastly, we perform competitive or even better than the state-of-the-art using box-supervision\footnote{Note that we only use the top-1 proposal per action per video, as more proposals per video skews AUC performance \citep{weinzaepfelICCV2015learningToTrack}}.

On UCF-101, we also outperform all existing weakly-supervised alternatives. Our approach reaches an mAP of 0.418, compared to 0.369 of \cite{li2018videolstm} and 0.351 of \cite{mettes2017localizing}, the state-of-the-art in weakly-supervised action localization. In comparison to box-supervision, we outperform the approach of \cite{gemert2015apt}, which employs identical spatio-temporal proposals and representations. On UCF-101, the state-of-the-art approaches in action localization from box-supervision perform better~\citep{kalogeiton2017action,yang2017spatio}. These approaches score and link 1 to 10 consecutive boxes into tubes, rather than opting for spatio-temporal proposals. Based on our second experiment, we posit that better spatio-temporal proposals can narrow this gap in performance.
%

Lastly, we provide results with our approach on Hollywood2Tubes. We first observe that overall, the performance on this dataset is lower than on UCF Sports and UCF-101 in terms of mAP scores. This indicates the challenging nature of the dataset. The combination of temporally untrimmed videos, multi-shot actions, and actions of complex semantic nature make for a difficult action localization.
%
%
Our approach provides a new state-of-the-art result in this dataset with an mAP of 0.178, compared to 0.143 of~\citep{mettes2016spot} and 0.172 of \citep{mettes2017localizing}.

\section{Conclusions}
\label{sec:conclusions}

This paper introduces point-supervision for action localization in videos. We start from spatio-temporal proposals, normally used during inference to determine the action location. We propose to bypass the need for box-supervision by learning directly from spatio-temporal proposals in training videos, guided by point-supervision. Experimental evaluation on three action localization datasets shows that our approach yields similar results to box-supervision. Moreover, our approach can handle sparse and noisy point annotations, resulting in a 20 to 150 times speed-up for action supervision. To help guide the selection of spatio-temporal proposals during inference, we propose pseudo-points, automatic visual cues in videos that hallucinate points in test videos. When weighted and selected properly with our quality measure, pseudo-points can have a positive impact on the action localization performance. We conclude that points provide a fast and viable alternative to boxes for spatio-temporal action localization.


\begin{appendices}
\section{Types of localization errors}
\label{sec:appendix}
For the error diagnosis, we consider five types of detections, parameterized by an overlap threshold $\tau$. The first detection type is a correct detection (detection from positive video with an overlap of at least $\tau$). The second type is a localization error (detection from positive video with an overlap less than $\tau$, but greater than 0.1). The third type is confusion with another action (detection from negative video with an overlap of at least 0.1). The fourth type is background detection from own action (detection from positive video with an overlap less than 0.1). The fifth and final type is background detection from another action (detection from negative video with an overlap less than 0.1). These five types cover all possible types of detections. These types are similar to~\citep{hoiem2012diagnosing}. Different from~\citep{hoiem2012diagnosing}, we do not split actions into similar and dissimilar (since no such subdivision exists). Instead, we split background detections into detection from own and other actions.
\end{appendices}

\section*{Acknowledgments}
This work is supported by the Intelligence Advanced Research Projects Activity (IARPA) via Department of Interior/Interior Business Center (DOI/IBC) contract number D17PC00343. The U.S. Government is authorized to reproduce and distribute reprints for Governmental purposes notwithstanding any copyright annotation thereon. Disclaimer: The views and conclusions contained herein are those of the authors and should not be interpreted as necessarily representing endorsements, either expressed or implied, of IARPA, DOI/IBC, or the U.S. Government.

\bibliographystyle{spbasic}      
\bibliography{point-supervision-ijcv}   

\end{document}